\pdfoutput=1
\documentclass[11pt]{article}

\usepackage[final]{acl} %review

\usepackage{times}
\usepackage{latexsym}
\usepackage{color, colortbl}

\definecolor{LightCyan}{rgb}{0.88,1,1}
\definecolor{lightgray}{rgb}{0.83, 0.83, 0.83}
\usepackage[T1]{fontenc}
\usepackage[utf8]{inputenc}

\usepackage{microtype}

\usepackage{inconsolata}

\usepackage{graphicx}

\usepackage{graphicx}               % Include graph
\usepackage{tabularx}               % Better table formatting
\usepackage{booktabs}
\usepackage{amsmath}
\usepackage{stmaryrd}
\usepackage{xcolor}
\usepackage{soul}

\newcolumntype{C}{>{\centering\arraybackslash}X}
\usepackage{multirow}               % Multi-row in tables
\usepackage{diagbox}                % Diagonal line in tables
\usepackage{hhline}                 % Draw double line in tables
\usepackage{color}                  % Text and background color
\usepackage{amsmath}                % Formula
\usepackage{amssymb}                % Formula
\usepackage{mathtools}              % Formula
\usepackage{enumitem}               % Better itemize environment

\usepackage{subfigure}
\usepackage{booktabs}
\usepackage{extarrows}
\usepackage{makecell}

\usepackage{soul}

\definecolor{RoseQuartzBg}{HTML}{F7CAC9}
\definecolor{RoseQuartz}{HTML}{F5A798}
\definecolor{Serenity}{HTML}{92A8D1}
\definecolor{OrangeRed}{rgb}{1.0, 0.27, 0.0}
\definecolor{Red}{rgb}{1.0, 0.0, 0.0}
\definecolor{Turquoise}{HTML}{0F4C81}
\definecolor{RoyalBlue}{cmyk}{1, 0.50, 0, 0}
\definecolor{Green}{rgb}{0.0, 1.0, 0.0}
\definecolor{Orchid}{rgb}{0.85, 0.44, 0.84}
\definecolor{Orange}{rgb}{1.0, 0.5, 0.0}
\usepackage{xparse}
\NewDocumentCommand{\lifu}{ mO{} }{\textcolor{Red}{\textsuperscript{\textit{Lifu}}\textsf{\textbf{\small[#1]}}}}
\NewDocumentCommand{\barry}{ mO{} }{\textcolor{blue}{\textsuperscript{\textit{Barry}}\textsf{\textbf{\small[#1]}}}}
\NewDocumentCommand{\hugo}{ mO{} }{\textcolor{Serenity}{\textsuperscript{\textit{Hugo}}\textsf{\textbf{\small[#1]}}}}
\NewDocumentCommand{\qf}{ mO{} }{\textcolor{RoseQuartzBg}{\textsuperscript{\textit{Qifan}}\textsf{\textbf{\small[#1]}}}}
\NewDocumentCommand{\sijia}{ mO{} }{\textcolor{RoyalBlue}{\textsuperscript{\textit{sijia}}\textsf{\textbf{\small[#1]}}}}
\NewDocumentCommand{\minqian}{ mO{} }{\textcolor{teal}{\textsuperscript{\textit{Minqian}}\textsf{\textbf{\small[#1]}}}}
\NewDocumentCommand{\zhiyang}{ mO{} }{\textcolor{OrangeRed}{\textsuperscript{\textit{Zhiyang}}\textsf{\textbf{\small[#1]}}}}

\usepackage{amssymb}% http://ctan.org/pkg/amssymb
\usepackage{pifont}% http://ctan.org/pkg/pifont

 \setlist[itemize]{leftmargin=*}
\setlist[enumerate]{leftmargin=*}

\newpage
\usepackage{appendix}  
\usepackage{diagbox}                % Diagonal line in tables

\usepackage{natbib} 
\usepackage{algorithm}
\usepackage{algorithmic}
 
\title{Error-driven Data-efficient Large Multimodal Model Tuning}

\author{Barry Menglong Yao \\
  UC Davis \\
  \texttt{bmyao@ucdavis.edu} \\\And
  Qifan Wang \\
  Meta AI \\
  \texttt{wqfcr@meta.com} \\\And 
  Lifu Huang \\
  UC Davis \\
  \texttt{lfuhuang@ucdavis.edu} \\}

\begin{document}
\maketitle

\begin{abstract}
% \barry{original paper title: Data-efficient Large Multimodal Model Tuning. The new title can better summarize the overall research question and approach.}
 %\st{However, their few-shot performance in real-world applications and downstream tasks remains suboptimal.}\hl{

Large Multimodal Models (LMMs) have demonstrated impressive performance across numerous academic benchmarks. However, fine-tuning still remains essential to achieve satisfactory performance on downstream tasks, while the task-specific tuning samples are usually not readily available or expensive and time-consuming to obtain. To address this, we propose an error-driven data-efficient tuning framework that aims to efficiently adapt generic LMMs to newly emerging tasks without requiring extensive task-specific training samples. In our approach, a generic LMM, acting as a student model, is first evaluated on a small validation set of the target task, and then a more powerful model, acting as a teacher model, identifies the erroneous steps within the student model's reasoning steps and analyzes its capability gaps from fully addressing the target task. Based on these gaps, targeted training samples are further retrieved from existing task-agnostic datasets to tune the student model and tailor it to the target task. We perform extensive experiments across three different training data scales and seven tasks, demonstrating that our training paradigm significantly and efficiently improves LMM's performance on downstream tasks, achieving an average performance boost of 10.3\%\footnote{The programs are publicly available at \url{https://github.com/PLUM-Lab/DELAMO_LMM_Tuning}.}. 
%\st{teacher model analyzes the weaknesses and missing skills of the student model based on its incorrect predictions. Our approach then retrieves targeted training data from a large-scale, domain-agnostic supporting dataset to address these specific weaknesses.} 
% \lifu{need to be carefully tuned, especially the approach description, to be aligned with the introduction} \barry{Done}
 % , achieving improvements of up to 4.82\% on MMBench, 4.17\% on ScienceQA, 2.10\% on Image-Caption Matching, 5.80\% on Visual Question Answering, 35.30\% on Appliance Classification, 23.30\% on Furniture Classification, and 38.20\% on Living Thing Classification over baseline methods.
%\st{COCO by 16.40\%, on Caltech by 11.30\%, and on MMBench by 6.48\%} \st{, even with limited training data, e.g., 3K training data.}\footnote{Our code and checkpoints will be made publicly available.}   
\end{abstract}
% Minqian: introduction task-specific improvement for LLMs 
% backbone -> test on different LLMs/MLMs, see what domains/tasks the LLMs are weak at, then improve. Whether it can be generalized to other MLMs
% baseline: only compared with random baseline; using sentence embeddings to retrieve training samples; 
% Figure 1: put step 1 to the top left; Figure 2: add 1, 2, 3,..; Prompt figures: better utilize the space
% 2.5 / 3

% Ying
% Baseline: other error driven approaches; 
% Efficiency comparison: multisteps v.s. one step
% Scope of introduction: differentiate it from continual learning; data-efficient fine-tuning instead of continual learning
% Add more qualitative examples 
% Figure 2 is not clear, add the correspondence of the probability to the reasoning steps;
% borderline

% Jingyuan
% name of skill is confusing

% Zihao
% gaps between abstract, introduction, and related work
% Figure 2 is not clear whether the four steps are fed as input to LLMs all together or separately

% Mo
% explain the drop of performance when increasing the number of samples in experiment table

% Zhiyang
% add more examples for skills; what do they look like? 
 
\section{Introduction}  

%what is the real-life problem

Pretrained large multimodal models (LMMs), such as GPT-4~\cite{achiam2023gpt} and LLaVA~\cite{liu2024visual}, have demonstrated %impressive capabilities and 
strong performance across various academic benchmark datasets~\cite{xu2022multiinstruct,reddy2022mumuqa,liu2024mmbench,lu2022learn,yue2024mmmumassivemultidisciplinemultimodal,yu2023mmvetevaluatinglargemultimodal}. 
%, which leads to their significant application in broad domains and downstream tasks, e.g., medical domain, financial domain, and law domain~\cite{}. 
However, when leveraging LMMs for real-world applications, despite direct task adaptation with techniques such as prompting~\cite{radford2019language,wei2023chainofthought,qi2023artsocraticquestioningrecursive,yao2024tree} or in-context learning~\cite{brown2020language,jiang2024many,zhao2024mmiclempoweringvisionlanguagemodel,doveh2024multimodalincontextlearningvision}, careful fine-tuning on a substantial amount of task-specific training samples is still essential in order to achieve satisfactory performance~\cite{luo2022biogpt,gu2021domain,liang-etal-2023-gpt,yao2023ameli}, while such task-specific training samples are usually not readily available or expensive and time-consuming to achieve. Therefore, a critical question that we would like to answer is: \textit{How can we effectively tune large multimodal models for newly emerging problems without requiring a large amount of task-specific training samples?}

One potential solution is to apply data augmentation methods to automatically synthesize or enlarge the training samples~\cite{lee2024llm2llm,10.48550/arxiv.2302.13007,li2024synthetic,zhao2024self,nayak2024learning,xu2023learning}. However, they usually lead to undesired effects, such as introducing significant \textit{bias} into the downstream tasks~\cite{angelakis2024data,lin2024good,muthukumar2020harmless,hastie2022surprises} or causing \textit{model collapse}~\cite{shumailov2023curse,feng2024tale}, where models tuned from synthesized training samples tend to forget the true underlying distribution of human-generated datasets. 
Additionally, several recent studies explored selecting relevant tasks or data samples from external resources to fine-tune the models for target tasks, where the selection is based on the similarity between the evaluation instances of the target task and training samples of other tasks using either features such as n-grams and task instructions~\cite{lee2024instruction,xie2023data,gururangan2020don} or gradients calculated from the model~\cite{xia2024less,han2023understanding}.
However, these approaches either necessitate a high degree of alignment between the surface forms of external datasets and the target task or rely on backward passes that are computationally intensive due to the large size of the external datasets. %} \lifu{add one sentence to explain why these approaches are not effective enough. Also, if you include this discussion, why don't you add these approaches as baselines?}

In this work, we propose a novel \textit{error-driven}, \textit{data-efficient} tuning paradigm that enables the effective adaptation of generic, pre-trained large multimodal models (LMMs) to diverse and emerging downstream tasks, while minimizing the need for extensive task-specific training samples. 
%to effectively adapt generic, pre-trained large multimodal models (LMMs) to various new and emerging downstream tasks without requiring a large amount of task-specific training samples. 
This paradigm is motivated by the \textit{gap detection and filling} process in human learning~\cite{bambrick2010driven}, where learners identify knowledge gaps and incrementally fill them through targeted exploration. Based on this motivation, we design a teacher-student framework where a pre-trained LMM, acting as the student model, is first applied to a small set of validation samples specific to the target task. The student model's predictions are then analyzed, and based on its errors, a teacher model—typically another large multimodal model (e.g., GPT-4o-mini)—is designed to identify the erroneous steps within the student model's reasoning processes, and further analyze and summarize its missing skills, representing the capability gaps preventing the student model from fully addressing the target task. After identifying these gaps, a set of tuning samples that are specifically related to the missing skills is retrieved from existing task-agnostic, large-scale supporting datasets, to fine-tune the student model.
To evaluate the effectiveness of our framework, we employ different student models, including 
LLaVA-7B~\cite{liu2024visual} and Qwen2-VL-7B~\cite{Qwen2VL}, and teacher models, including GPT-4o-mini~\cite{achiam2023gpt} and LLaVA-OneVision-72B~\cite{li2024llava},
%as the student model, GPT-4o-mini~\cite{achiam2023gpt} as the teacher model, 
and
conduct extensive experiments across seven tasks and datasets, including MM-Bench~\cite{liu2024mmbench}, a comprehensive benchmark covering a wide range of multimodal processing tasks, and six downstream tasks including {ScienceQA}~\cite{lu2022learn}, %and utilized to showcase the potential of our framework as a post-pretraining step to further enhance the general capabilities of pre-trained LMMs, \hl{ScienceQA}~\cite{lu2022learn}, \hl{which is a comprehensive collection of multimodal science questions and answers, designed to facilitate research in understanding and reasoning across text, images, and diagrams in scientific contexts, }and five downstream tasks, including  
\text{Appliance Classification}~\cite{lin2014microsoft}, \text{Furniture Classification}~\cite{lin2014microsoft}, \text{Living Thing Classification}~\cite{li_andreeto_ranzato_perona_2022}, \text{Vision Question Answering}~\cite{zhu2016visual7w}, and \text{Image Caption Match}~\cite{lin2014microsoft}. 
We utilize Vision-Flan~\cite{xu2024vision} as the external supporting dataset as it covers hundreds of existing human-labeled tasks and datasets. Across different numbers of tuning samples retrieved from the supporting dataset, our approach significantly outperforms other data selection baselines as well as the LMM that is fine-tuned on the whole supporting dataset, 
%an average performance boost of 7.01\% over the pre-trained LMM, 
highlighting the efficiency and effectiveness of our error-driven, data-efficient tuning framework in task adaptation.
Our contributions are summarized as follows:

%FIXME\qf{Better to articulate at least three points of contributions. Consider split the first bullet point into two.}
 
\begin{itemize}

% \item \hl{ We propose a straightforward yet effective method to identify the erroneous steps within LMM's rationales}
\item We propose a novel error-driven, data-efficient tuning framework that identifies capability gaps in LMMs
%'s (LMM) capabilities on new downstream tasks 
and retrieves targeted tuning samples from existing datasets to effectively adapt them to new downstream tasks without 
%to address these deficiencies, 
%thereby improving the LMM's performance without 
requiring extensive task-specific training samples.
 
% \vspace{-2mm}

\item We conduct comprehensive experiments, demonstrating that our framework significantly surpasses all baseline methods in effectively adapting generic LMMs to specific downstream tasks %without requiring task-specific training samples.
%with substantial performance gains 
while incurring minimal training costs.

\end{itemize}

\section{Related Work}  
%training strategy: data selection, curriculum learning, knowledge distillation
\paragraph{Error-driven Learning}

Inspired by cognitive science, error-driven learning~\cite{carpenter1987massively,hoppe2022exploration} enhances model performance by updating parameters based on error samples~\cite{rumelhart1986learning} or explicitly analyzing and addressing errors. %~\cite{10.48550/arxiv.2310.15746,wang2023learning,10.48550/arxiv.2305.08844,WendaXu2023}. 
%emerges as a new paradigm to boost model performance, by either directly updating model parameters based on the loss computed on the error samples~\cite{rumelhart1986learning} or explicitly analyzing errors and addressing them through various modules~\cite{10.48550/arxiv.2310.15746,wang2023learning,10.48550/arxiv.2305.08844,WendaXu2023}. 
For instance,
~\newcite{10.48550/arxiv.2310.15746} and~\newcite{wang2023learning} directly prompt large language models (LLMs) to summarize error-driven guidance and integrate it into subsequent prompts. 
%summarize guidance from a set of error samples and append the guidance into the prompt to avoid making similar errors on subsequence data samples. 
%\hl{given error samples, previous studies}\cite{10.48550/arxiv.2310.15746, wang2023learning}\lifu{use newcite when the references are used as subjects}\barry{Done} \hl{prompt LLMs to summarize the guidances to avoid these errors, e.g., }``\hl{If the review contains profanity or vulgar language, then it may be offensive depending on the context and severity}''. \hl{These guidances are then appended into the prompt of the LLMs to avoid making similar mistakes when processing subsequent inputs}. \st{summarize the errors as guidance to refine the following prediction}\lifu{not clear enough... how the errors are used as guidance to refine the predictions?}\barry{Done}. 
\newcite{10.48550/arxiv.2305.08844} and \newcite{WendaXu2023} introduce critique generators to refine predictions during inference. 
%propose a critique generator to pinpoint defects within the current prediction so that LLMs can refine the prediction based on the critique during inference.  %\st{to analyze errors to generate feedback that improves the current predictions}\lifu{again not clear enough}\barry{Done}. 
Other studies~\cite{lee2024llm2llm, Shengnan2023a, 10.48550/arxiv.2305.03610, 10.48550/arxiv.2310.18628,wang2024targeted} %\st{works}\lifu{work is not countable when it's referring to research work. Use studies instead}\barry{Done}\textcolor{teal}{\st{, such as}}\lifu{remove it}\barry{Done} \cite{lee2024llm2llm, Shengnan2023a, 10.48550/arxiv.2305.03610, 10.48550/arxiv.2310.18628,wang2024targeted}\lifu{add Sijia's recent work for targeted augmentation}\barry{Done. added}, 
propose targeted data augmentation that automatically generates synthetic data using error samples. 
Unlike these methods, our approach fine-tunes LMMs by retrieving training samples from large-scale, domain-agnostic datasets, addressing missing skills identified from error samples.
%which automatically generates synthetic data based on these error samples, instead of all seed data. 
% In contrast to all these studies, our approach conducts further fine-tuning of LMMs based on the additional training samples retrieved from existing and domain-agnostic large-scale datasets, driven by the missing skills of LMMs analyzed from the error samples.

%targeted LMM tuning by retrieving targeted training data from ever-increasing domain-agnostic large-scale datasets for subsequent training iterations.
%\st{With the increasing availability of domain-agnostic large-scale datasets}\lifu{it's debatable whether the datasets are domain-agnostic because they just contain many datasets for different subtasks but each dataset is still domain specific}, such as Vision-Flan~\cite{xu2024vision}\lifu{add reference}\barry{Done}, \st{we propose analyzing errors and addressing them by selecting targeted training data for subsequent training iterations.}\lifu{not clear the difference between our work and previous studies or the limitations of previous studies}\barry{added description about the difference}

\begin{figure*}[hbt!]
    \centering
    \includegraphics[width=0.85\textwidth]{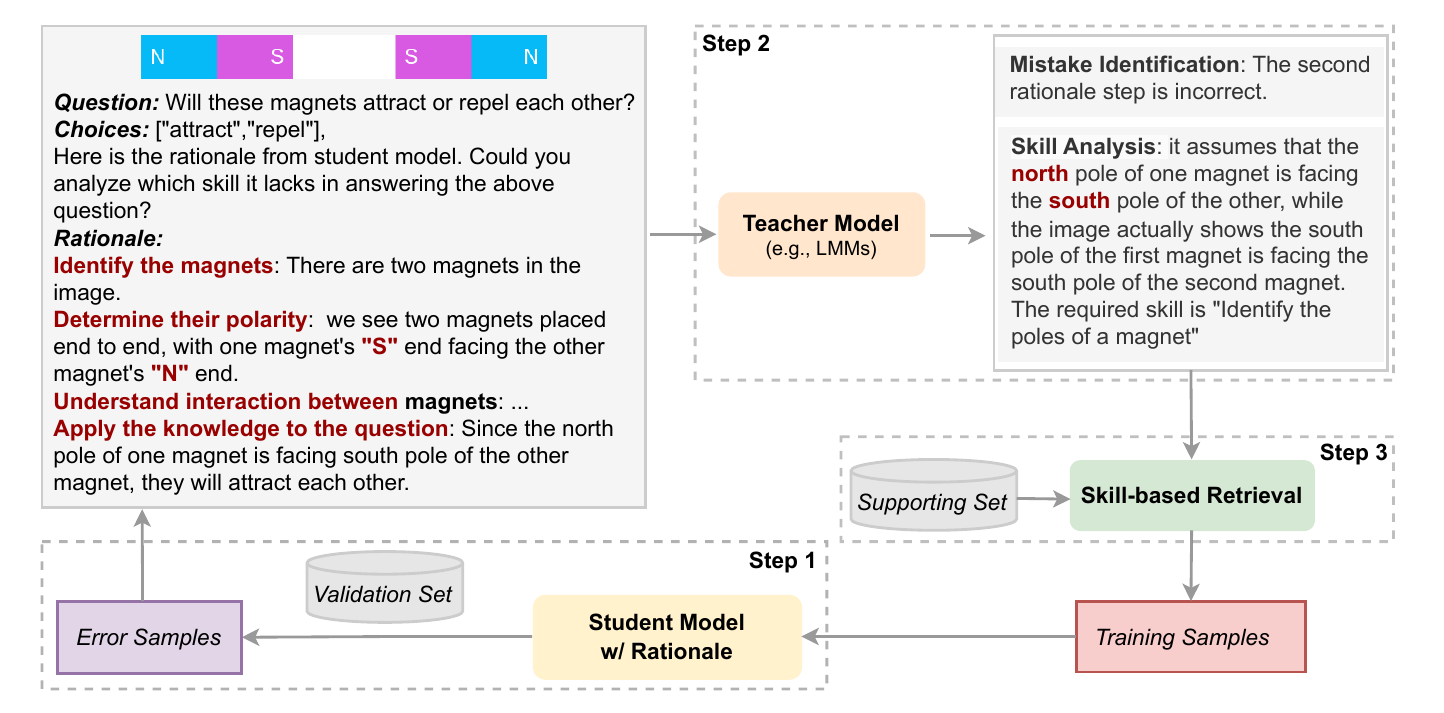}
    \vspace{-4mm}
    \caption{Overview of the error-driven data-efficient tuning paradigm. %\lifu{make the left text box larger; reduce the space in the right text box; Add a little margin between the text and the border of the box; Make text to be in the center of each model based block, e.g., Training samples}\barry{Done} %\lifu{make the text boxes wider so it fits better to the two-column width and saves more space}\barry{Done} %In \textbf{Step 1}, a student model (i.e., a pre-trained LMM) is first evaluated on the validation dataset to obtain error samples and corresponding rationales. In \textbf{Step 2}, based on the wrong predictions and rationales from the student model,  a teacher model (i.e., GPT-4o-mini) identifies the mistakes from the rationales and analyzes the missing skills from the current student model. In \textbf{Step 3}, based on missing skills of the student model, we retrieve a set of targeted training samples from an external supporting dataset to fine-tune the student model.%\lifu{make the steps better aligned with the corresponding blocks based on the paper discussion}\barry{Done}
    }
    \label{fig:pipeline}
    \vspace{-5mm}
\end{figure*}

% \vspace{-2mm}
\paragraph{Data Selection}
Data selection is often framed as a coreset selection problem \cite{10.48550/arxiv.1601.00617}, aiming to identify a subset of training examples that achieves comparable performance to the full dataset. This is typically done by assessing training data quality \cite{10.48550/arxiv.2403.09559,chen2023alpagasus,zhou2024lima,10.48550/arxiv.1812.05159, sener2017active, 10.48550/arxiv.2103.00123,Mengzhou_2024} or selecting high-uncertainty samples \cite{10.48550/arxiv.2311.00288,liu2024selectit}. %to select the most valuable training samples. 
%Within this broader domain, t
Targeted data selection refines this approach by choosing fine-tuning data aligned with the target distribution, using similarity measures based on surface features~\cite{lee2024instruction,xie2023data,gururangan2020don} or LLM gradient vectors~\cite{xia2024less,han2023understanding}. Unlike these methods, our approach directly identifies LMM weaknesses from error samples, enabling more data-efficient and computation-efficient sample selection.

\paragraph{Curriculum Learning}
%Inspired by the cognitive principle of humankind, \newcite{10.1145/1553374.1553380} firstly introduced the Curriculum Learning (CL) paradigm where a model is trained in a meaningful order, from the easy samples to the hard ones. 
Curriculum Learning (CL) trains models in a structured order, progressing from easy to hard samples. 
Early CL studies~\cite{10.1145/1553374.1553380,spitkovsky2010baby} relied on rule-based criteria (e.g., training on shorter sequences first). 
%leverage rule-based criterion for sample selection, e.g., longer sequences are harder to predict than shorter ones so the model is gradually trained from shorter samples to longer ones. 
Self-paced learning methods~\cite{kumar2010self,lee2011learning,ma2017self} dynamically select samples based on model performance, training loss, or likelihood. 
%compute hardness for all the samples based on the model learning dynamics, such as model performance, training loss, likelihood of predictions, and so on.  
%Several recent studies~\cite{10.48550/arxiv.1707.00183,kim2018screenernet,hacohen2019power,zhang2019automatic} proposed teacher-student architectures where a teacher model is designed to select suitable training samples for a student model based on a reinforcement learning reward. 
Recent teacher-student approaches~\cite{10.48550/arxiv.1707.00183,kim2018screenernet,hacohen2019power,zhang2019automatic} use reinforcement learning to guide selection. 
Our method extends this framework by introducing Mistake Identification and Skill Analysis to efficiently detect and address the student model’s weaknesses. 
%Our work shares a similar teacher-student framework, but different from all these prior studies, we introduce two novel components, including Mistake Identification and Skill Analysis, to efficiently identify the weakness of the student model driven by the errors it has made.

%A detailed related work discussion can be found in Appendix ~\ref{sec:related_work}.\lifu{don't put related work to appendix. First try to cut it}
%and address the student model's weaknesses in an error-driven manner 
%Teacher-Student Curriculum Learning and refine our curriculum in an error-driven manner. 
%\lifu{revise }\barry{Done}

% \paragraph{Task Adaptation}
% \paragraph{Mistake Identification}, like Big-Mistake
% \paragraph{Skill Analysis}, like Skill-it, latent skill, MM-Bench

% \paragraph{Knowledge Distillation}, somehow related, but not significantly relevant

\section{Approach}   
\subsection{Overview}  
%\lifu{need to better formulate the problem, e.g., no one knows what the student model means, how it is used, what are the datasets used for, what's the goal of the problem, etc.} \barry{Done. Changed student model to LMMs. Stress out the goal. As for the dataset, the training paradigm should work for any given datasets and downstream tasks, so we specify the used datasets in the Experiment section instead of in the Approach section}
%\lifu{refine the mathematical notations. The current ones look too many subscripts or the subscripts are too long}\barry{Done. refined the notations. Note that notations like $\mathcal{D}_{train}$ and $\mathcal{D}_{val}$ are also used in previous work }
%\lifu{find some previous papers and improve the description of the overview. The current paragraphs read not neat enough}

    % refer to related work
    % 1.Lee, N. et al. LLM2LLM: Boosting LLMs with Novel Iterative Data Enhancement. arXiv (2024).
    % 2.Learning from Mistakes via Cooperative Study Assistant for Large Language Models. (2023).
    % 3.Yang, Z., Li, P. & Liu, Y. Failures Pave the Way: Enhancing Large Language Models through Tuning-free Rule Accumulation. arXiv (2023) doi:10.48550/arxiv.2310.15746.
    % 4.An, S. et al. Learning From Mistakes Makes LLM Better Reasoner. arXiv (2023).

Given a new task with a test set $\mathcal{D}_{test}$ and a validation set $\mathcal{D}_{val}$, we aim to efficiently adapt a generic, pre-trained large multimodal model (LMM) to it without requiring extensive task-specific training samples. To achieve this, we propose an error-driven data-efficient tuning framework, as shown in Figure~\ref{fig:pipeline}, consisting of three iterative steps: Step 1 (\textbf{Error Collection}) identifies error samples by evaluating the student model's predictions and rationales on validation samples. Step 2 (\textbf{Mistake Identification and Skill Analysis}) uses a teacher model to pinpoint the key erroneous step and infer the missing skill needed for improvement. Note that while most downstream tasks require diverse skills, a pre-trained LLM may have already excelled in some, so we mainly focus on identifying and enhancing the missing skills in the given LMM. Step 3 (\textbf{Targeted Tuning}) further retrieves targeted samples from existing datasets to fine-tune the student model, refining its capabilities for the missing skills. These three steps iterate until the maximum number of iterations is reached. In the following, we provide details for each component.

 \begin{figure*}[!ht] 
  \centering
    \includegraphics[width=0.9\textwidth]{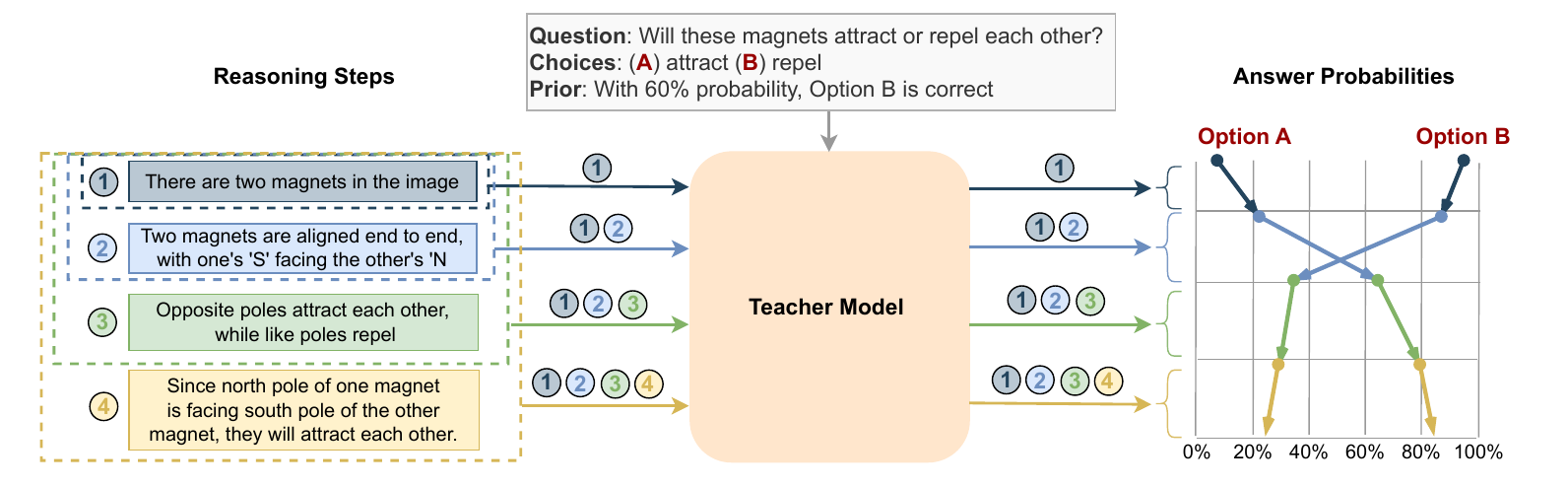}
    \vspace{-5mm}
    \caption{Example for illustrating the process of mistake identification. At each iteration, we append one more reasoning step into the prompt to ask the teacher model to answer the question and track the probability changes of all the candidate option tokens.  \iffalse  \lifu{make the figure centralized in the paper}\barry{Done}\fi %\lifu{make the middle box wider, so the Question could be one line; make the left and right subfigures wider and shorter to save more space}\barry{Done}%\lifu{make the middle box wider to save more space}\barry{Done. changed text from 'There is a probability of 60\% that option B is correct' to 'With 60\% probability, Option B is correct' }
    }
    %FIXME \qf{Can we make the Question text box on the top of the figure wider and short} }
    \vspace{-6mm}
    \label{fig:mistake_identification}
\end{figure*}

\subsection{Error Collection from Student Model}\label{sec:error_extraction}
Given a target task with a validation set $\mathcal{D}_{val}$, we leverage a generic and pre-trained LMM as the student model $\mathcal{M}_{S}$, which is prompted to generate a sequence of intermediate reasoning steps~\cite{wei2023chainofthought} and a final answer for each validation sample. The intermediate reasoning steps are viewed as a rationale for the predicted answer. 
The LMM is prompted to specifically follow an answer format such as ``The final answer is option A'', and we will directly parse the final answer from the model's response based on the answer format.\footnote{We also consider the variants of the answer format shown in Table~\ref{tab:answer_format} in Appendix ~\ref{sec:answer_format}.} An example prompt for ScienceQA task is shown in Figure~\ref{fig:inference} in Appendix \ref{sec:inference_prompt}. We finally compare the predicted answer with gold answer for each validation example and obtain a set of error samples and their corresponding intermediate reasoning steps as rationales.

%\hl{Given the generic and pre-trained LMM and the validation set, the goal of \textbf{Error Sample Extraction} module is to extract validation samples where LMM predicts wrongly, along with the corresponding rationales from LMM. We leverage the prompt template shown in Figure}~\ref{fig:inference} \hl{to trigger LMM to first generate step-by-step rationales and finally provide the answer for each validation sample in the} Chain-of-Thought~\cite{wei2023chainofthought} manner. Following previous work~\cite{liu2024visual}, \hl{we specify the required answer format like ``The final answer is option A'' in our prompt template and then parse the answer from the LMM's response based on the answer format and its variants as shown in Table}~\ref{tab:answer_format} in Appendix. \hl{ We finally compare predicted option with gold option to obtain the error samples and their corresponding rationales}. 

\subsection{Mistake Identification}\label{sec:mistake_identify} 

Given an error sample containing a question $q$, a wrong prediction $y$ with a rationale $r$ from the student model, and a gold answer $\tilde{y}$, we first split the rationale $r$ into a sequence of reasoning steps $r=[r_1,r_2,...]$. Here, we follow previous study~\cite{tyen2024llms} and treat each sentence in the rationale as one reasoning step. The goal of \textbf{Mistake Identification} is to apply a new large multimodal model, e.g., GPT-4o-mini~\cite{achiam2023gpt}, as a teacher model $\mathcal{M}_{T}$ to analyze the incorrect predictions from the student model, and  
locate the \textit{Mistake Step} $r_m$, a.k.a., the most significant erroneous reasoning step that leads to the final incorrect answer, from the rationale. Motivated by previous studies~\cite{tyen2024llms}, we define the most significant erroneous reasoning step $r_m$ as the first rationale step that leads to the prediction of the wrong answer $y$. For example, for the error sample shown in Figure \ref{fig:pipeline}, the second step ``\textit{one magnet's south end facing the other magnet's north end}'' is identified as the mistake step as it contributes most to the final wrong prediction of the student model.

We propose an \textit{answer-switch} based method to identify the mistake step, as shown in Figure~\ref{fig:mistake_identification}. The core idea is to prompt the teacher model to respond to the same question using the rationales provided by the student model. We then analyze the changes in the probabilities of the candidate answers as each individual rationale step is incrementally appended. To encourage the teacher model to favor the correct answer at the beginning, we modify the prompt to include prior knowledge that indicates a higher probability for the correct answer, e.g., ``\textit{There is a probability of 60\% that option B (repel) is correct}'', and instruct the teacher model to rely on this prior knowledge if it lacks sufficient information to determine the answer. We then gradually append each reasoning step into the prompt of the teacher model $\mathcal{M}_{T}$ and monitor the changes in the model prediction, with the expectation that the probability of wrong answer $y$ will gradually become higher after we append the erroneous reasoning steps. 

Our preliminary experiments on ScienceQA with 100 error samples have shown that when the teacher model is provided with the correct answer as prior knowledge, it initially could assign a higher probability to the correct answer in 89\% of cases. As the student model's incorrect reasoning steps are added, the teacher model shifts to the wrong answer in 70\% of cases. Without prior knowledge, the correct answer receives a higher initial probability in 60\% of cases, with a shift to the wrong answer in 43\%.
These results support the design of our \textit{answer-switch} based approach for mistake identification. In addition, we also restrict the teacher model from accessing the image so that it's forced to choose the answer solely based on the reasoning steps of the student model. %\lifu{no one will put such as long paragraph as a footnote. Add more details to make this paragraph and the experiments clearer}\barry{Maybe we can add this preliminary experiments into appendix instead of adding too many experiment details in Approach section \ref{sec:answer_switch_experiment}}

Figure~\ref{fig:mistake_identification_prompt} in Appendix~\ref{sec:mistake_prompt} shows the prompt template for mistake identification. For each round of inference, the input prompt to $\mathcal{M}_{T}$ consists of the question ``\textit{Will these magnets attract or repel each other?}'', the prior knowledge about the correct answer ``\textit{There is a probability of 60\% that option B is correct}'', and a subset of reasoning steps, while the output consists of a template-based answer, e.g., ``\textit{The answer is the option A}''.\footnote{For non-multiple-choice tasks, we convert them by setting the gold answer and the wrong prediction as two options.} To determine the probability of each candidate option, we first identify the position of the option token (e.g., ``A'') in the answer, %e.g., ``A'', within the generated tokens, e.g., ``The answer is the option A''. At this token position, we 
and obtain the probabilities of other candidate option tokens such as ``B'', ``C'', and ``D'', from the teacher model. 
This process is repeated as we sequentially append each reasoning step to the prompt, enabling us to track the probabilities of all answer options across iterations, e.g., $\textstyle{\{P(A|q,r_1),P(A|q,r_1,r_2),...,P(A|q,r_1,...,r_{i})\}}$, $\textstyle{\{P(B|q,r_1),P(B|q,r_1,r_2),...,P(B|q,r_1,...,r_{i})\}}$, respectively. Based on the change in probabilities of the correct answer ``B'' and the wrong answer ``A'', we identify the mistake step $r_m$ as the first reasoning step that causes the probability of the wrong answer to be higher than the probability of the correct answer by a predefined margin $\delta$ and the margin is maintained for the following $\lambda$ iterations:
\resizebox{1.02\columnwidth}{!}{
\begin{minipage}{1.02\columnwidth}
\vspace{-1mm}
\begin{align*}
m &:= \min \left\{ i \mid \forall j \in \{ 0, \ldots, \lambda-1 \}, \right.\\
    & \quad \left. P(A \mid q, r_1, \ldots, r_{i+j}) - \delta \ge P(B \mid q, r_1, \ldots, r_{i+j}) \right\} 
\end{align*}
\vspace{-1mm}
\end{minipage}
}
where $\delta$ is the probability gap between the wrong answer and the correct answer, and $\lambda$ is the number of steps where the probability gap persists.\footnote{We manually labeled the mistake step for 100 ScienceQA validation error examples and tuned $\delta$ and $\lambda$ on them (see results and probability gap statistics in Appendix~\ref{sec:mistake_hyper_search} and Section~\ref{sec:mistake_identification}).} %\lifu{this approach is not aligned with the definition of the mistake step in the previous paragraph --- it's not just the first step that leads to the wrong prediction, but also require a probability gap for several steps} \barry{1. The definition of mistake step is the first reasoning step which leads to the wrong prediction. Our data annotation also follows this definition. 2. There exists variance in the softmax probabilities from LLM. For the regularization purpose, we implement it with probability gap for several steps.}

\subsection{Skill Analysis}

After identifying the erroneous reasoning step $r_m$ from the rationale of each error sample, we further perform \textbf{Skill Analysis}, where the teacher model $\mathcal{M}_{T}$ is prompted to summarize one missing skill $s$,\footnote{
We follow ~\cite{10.48550/arxiv.2307.14430} and define a skill as a unit of behavior with associated data $X$ such that if the LMM is trained on dataset $D$, where $D \subseteq X$, it has improved performance on samples belonging to $X\backslash D$. See Appendix~\ref{sec:skill_definition} for more details on skill definition.
} 
% \footnote{We follow the definition of \textit{skill} in previous work ~\cite{10.48550/arxiv.2307.14430}, a.k.a., a skill $s$ is a unit of behavior with associated data $\mathcal{D}_s\subseteq \mathcal{D}$ such that if the LMM is trained on dataset $\mathcal{D}_{train}\subset\mathcal{D}_s$, then LMM has improved performance on samples belonging to $\mathcal{D}_{train}\backslash\mathcal{D}_{s}$ on average.}
such as \textit{identifying the poles of a magnet} in Figure~\ref{fig:pipeline}, which is required to correct the wrong reasoning step $r_m$.  
%\st{a skill $s$ as a unit of behavior with associated data $\mathcal{D}_s$ such that if the LMM is trained on $\mathcal{D}_s$, it has improved performance on samples belonging to $\mathcal{D}_{s}$.}\barry{the previous definition is oversimplified}\hl{  
Note that, for each error sample, we focus on one missing skill in one iteration and leave other missing skills for the following iterations. 
To achieve this goal, we design an in-context learning (ICL)~\cite{wei2022emergent,wei2022chain} based approach where the input of each in-context exemplar consists of a question together with its correct answer, complete rationale steps and a mistake step, and the output is the missing skill which is required to correct the mistake. %We leverage these in-context exemplars to guide the teacher model to generate a natural language explanation about the reason of the mistake step and summarize one skill to prevent this mistake for the following prediction. 
The prompt template for \textbf{Skill Analysis} is shown in Figure~\ref{fig:skill_analysis} in Appendix \ref{sec:skill_prompt}.

\begin{table*}[hbt!]%{r}{0.45\textwidth}%[h]
\small
\centering
\resizebox{1\textwidth}{!}{%
\begin{tabular}{l|c|c| c|c|c|c|c|c}
\toprule
\textbf{Method}&\textbf{\# of Tuning Samples} &\textbf{MM-Bench}   &\textbf{Appliance Cls}  &\textbf{Furniture Cls} &\textbf{Living Thing Cls} &\textbf{VQA}&\textbf{Image-Cap Match}  &\textbf{ScienceQA}  \\ % &

\midrule
Pre-trained LMM & 0 & 64.30$^\dagger$ &45.80 &  49.00&  79.40& 77.00&  64.10 &65.34\\ 
\midrule
  
Random&10K  & 62.85 & 57.47 &60.60 &  82.10 &74.03  &65.03  & 63.66     \\  
Superfiltering&10K & 62.65& 49.90&53.60 &77.00 & 75.30&68.40 &  64.15\\
INSTA*&10K & 63.25  & 60.00&  64.10& 89.20 &  72.20 &74.70   &62.52  \\  
\rowcolor{lightgray}
Our Approach &10K & \textbf{63.86}  & \textbf{62.10}& \textbf{64.80} &  \textbf{90.60} &  \textbf{76.00}  & \textbf{77.70}  &\textbf{65.89}  \\  
\midrule
Random&30K  & 62.60 &61.07  &63.13 & 86.83  &75.50  &71.97  & 63.38     \\ 
  Superfiltering&30K & 62.95& 53.40& 53.80& 78.40&76.90 & 69.90&65.05  \\
INSTA*&30K & 63.25 &61.90 &66.10  &  92.90&  72.10 & 76.90   & 65.39 \\   
\rowcolor{lightgray}
Our Approach&30K & \textbf{64.01} & \textbf{62.20}& \textbf{67.10} & \textbf{93.30} &  \textbf{77.30} &  \textbf{80.00}  &  \textbf{67.53} \\  

\midrule
Random&100K  & 62.95 &  60.83& 66.23& 88.67  & 76.90 & 77.50 & 64.55     \\
 Superfiltering&100K & 63.25& 55.60&54.90 & 82.90& 77.40& 73.50&65.00  \\
INSTA*&100K &62.05  & 62.90& 66.80 & 92.80 &74.00   & 77.60 &  65.25   \\  
\rowcolor{lightgray}
Our Approach&100K &  \textbf{64.46} & \textbf{64.10}& \textbf{67.70} & \textbf{93.60} & \textbf{79.00}   & \textbf{80.10}  & \textbf{68.02}   \\    
\midrule
Full Data &  1,552K &60.24 &63.50 & 69.80 &90.60  & 74.90 & 84.70  & 67.23 \\
 
\midrule
Validation Data &  1K & 63.86 & 59.90 & 57.80&89.00&77.40&67.80 & 65.39\\

\bottomrule
\end{tabular}
}
\vspace{-2mm}
\caption{Evaluation results on seven downstream tasks with different numbers of tuning samples retrieved from the supporting dataset (\%). \textbf{Full Data} means that the whole supporting dataset is used to tune the LMM while \textbf{Validation Data} stands for fine-tuning the pre-trained LMM on 1K validation samples of the target task. $^\dagger$Reported by \citet{liu2024improved}.
}
\label{tab:main_experiment_llava}
\vspace{-6mm}
\end{table*}

\subsection{Targeted Tuning}  

After analyzing the missing skills for all the error samples from the validation set $\mathcal{D}_{val}$, we then retrieve a set of relevant training samples from a domain-agnostic large-scale supporting dataset~\footnote{Our framework does not require the supporting dataset to be semantically similar to the downstream tasks. Instead, it capitalizes on underlying skills—such as counting and spatial relation recognition—that are shared between target tasks and existing task-agnostic instruction-tuning datasets.} to construct a targeted tuning dataset $\mathcal{D}_{train}$
and utilize $\mathcal{D}_{train}$ to fine-tune the student model to enhance its capability and address the identified skill gaps for the target downstream task. 

Specifically, %we leverage the existing domain-agnostic large-scale multimodal benchmark datasets such as LLaVA-665k and Vision-Flan-1-million as the supporting set $\mathcal{D}$, and 
for each sample in the supporting dataset, we pre-compute a set of required skills by prompting the teacher model to follow in-context exemplars and provide detailed analysis of the skills that are required to achieve the correct answer. The prompt template is shown in Figure~\ref{fig:skill_set_analysis} in Appendix~\ref{sec:skill_set_prompt}.  Then, for each error sample in $\mathcal{D}_{val}$, we apply BM25~\cite{robertson2009probabilistic} to calculate similarity scores between its missing skill $s$ and the concatenation of all required skills of each sample in the supporting dataset. The samples in the supporting dataset are then ranked according to the similarity scores, and the top-$K$ samples are selected as the training samples to improve the missing skills of the student model.

\section{Experiment}\label{sec:experiment} 
% We conduct experiments on \cite{li2023object} dataset for the attribute prediction task, and compare with curriculum learning baselines \cite{10.18653/v1/2022.naacl-main.72,10.48550/arxiv.1707.00183,zhang2022progressive}

%  \end{enumerate}

% \input{table/1ScienceQA}
% Table. \ref{tab:sqa}

% Table. \ref{tab:sqa_report}

\subsection{Experimental Setup}
For evaluation, we experiment with two different student models, including the instruction-tuned LLaVA-v1.5-7B~\cite{liu2024visual}\footnote{\url{https://huggingface.co/liuhaotian/llava-v1.5-7b}} and Qwen2-VL-7B~\cite{Qwen2VL,Qwen-VL}\footnote{\url{https://huggingface.co/Qwen/Qwen2-VL-7B-Instruct}}, and two different teacher models, including GPT-4o-mini~\cite{achiam2023gpt} (\texttt{gpt-4o-mini-2024-07-18}) and  LLaVA-OneVision-72B~\cite{li2024llava}\footnote{\url{https://huggingface.co/lmms-lab/llava-onevision-qwen2-72b-ov-chat}}, 
%first use the instruction-tuned LLaVA-v1.5-7B~\cite{liu2024visual}\footnote{\url{https://huggingface.co/liuhaotian/llava-v1.5-7b}} as the student model and GPT-4o-mini~\cite{achiam2023gpt}\footnote{We use the gpt-4o-mini-2024-07-18 version.} as the teacher model, %\lifu{how about mistake identification?}\barry{ the goal of teacher model is to strictly follow student model's rationale step and switch to the wrong answer when the wrong rationale step is fed. We can then identify the wrong rationale step based on this answer-switch phenomenon. So we do not need to feed image caption into teacher model for Mistake Identification}. 
%We then 
and evaluate our framework on seven downstream tasks and datasets: \textbf{MM-Bench}, a generic benchmark dataset for evaluating large multimodal models and covering diverse categories of tasks such as \textit{Attribute Recognition}, \textit{Action Recognition}, \textit{Object Localization}, and so on. MM-Bench is used to demonstrate the potential of our error-driven efficient-tuning framework as a post pre-training step to further improve the general capabilities of large multimodal models; and
%\textbf{ScienceQA}~\cite{lu2022learn}, 
%which is a comprehensive collection of multimodal science questions and answers, designed to facilitate research in understanding and reasoning across text, images, and diagrams in scientific contexts, and 
six downstream tasks, including \textbf{ScienceQA}~\cite{lu2022learn}, \textbf{Appliance Classification}~\cite{lin2014microsoft}, \textbf{Furniture Classification}~\cite{lin2014microsoft}, \textbf{Living Thing Classification}~\cite{li_andreeto_ranzato_perona_2022}, \textbf{Vision Question Answering}~\cite{zhu2016visual7w}, and \textbf{Image Caption Match}~\cite{lin2014microsoft}. %, that are sampled from the Vision-Flan-1-million dataset~\cite{xu2024vision}\lifu{add}\barry{Done}. 
For ScienceQA and MM-Bench, we use their official test sets. For each of the other downstream tasks, we sample $1$K data points as the test set and $1$K data points as the validation set. These tasks are employed to demonstrate the efficiency of our framework in adapting the generic pre-trained large multimodal model to specific downstream tasks. We use 
%To demonstrate the robustness of our framework, %on all the seven evaluation tasks, 
%we leverage the domain-agnostic large-scale dataset as the supporting dataset to retrieve targeted tuning samples:  
\textbf{Vision-Flan-1-million}~\cite{xu2024vision}\footnote{We removed all samples related to the seven evaluation tasks in Vision-Flan-1-million to ensure no overlap.} 
%We removed all the samples of the seven evaluation tasks from Vision-Flan-1-million 
%Since we sampled five tasks from Vision-Flan-1-million as the evaluation datasets, we filter all data points with the same task name in Vision-Flan-1-million 
%to guarantee that there is no overlap between the evaluation dataset and the supporting dataset.} 
as the supporting dataset as it covers hundreds of existing tasks and datasets created by humans. %To demonstrate the generalizability of our framework, we also experiment with 

We compare the student model tuned using our error-driven data-efficient tuning framework with four baselines: (1) \textbf{Pre-trained LMM}, which denotes the vanilla student model without any tuning; %(2) \textbf{Validation}, which uses the validation set to tune LMM; 
(2) \textbf{Random Sampling}, where the training samples are randomly sampled from the supporting set. This process was repeated three times, and the average performance is reported in Table~\ref{tab:main_experiment_llava}. Detailed results for each run are reported in Table~\ref{tab:random_sampling_result} in Appendix~\ref{sec:random_sampling}. (3) \textbf{INSTA*}~\cite{lee2024instruction}, which ranks %is adapted to our setting to rank 
the training samples based on their SBERT~\cite{reimers-2019-sentence-bert} similarity scores to the %between the training instances and 
validation samples, 
and select the same number of samples for targeted tuning. and (4) \textbf{Superfiltering}~\cite{li2024superfiltering}, which utilizes a small GPT-2 model~\cite{radford2019language} to filter out the high-quality subset based on instruction-following difficulty (IFD) score~\cite{li2023quantity}. Additionally, to better demonstrate the effectiveness and efficiency of our error-driven model tuning framework,  we also show the performance of the student model that is fine-tuned on the whole supporting dataset (\textbf{Full Data}) or the 1K task-specific validation samples (\textbf{Validation Data}).  %\barry{Plan to move baseline introductions to Appendix due to space limitation} 

\subsection{Main Results}~\label{sec:main_experiment}
Table~\ref{tab:main_experiment_llava} shows the performance of our framework with LLaVA-v1.5-7B as the student model and GPT-4o-mini as the teacher model, using different numbers of tuning samples from the supporting dataset. We compare our approach with several baselines and can see that: 
%results of both our approach and the baselines when using different numbers of tuning samples retrieved from the supporting dataset—Vision-Flan-1-million. 
%We can see that: 
(1) The pre-trained LMM underperforms on some tasks such as Appliance Classification (45.80\% accuracy) and Furniture Classification (49.00\% accuracy), 
%, e.g., 45.80\% accuracy on Appliance Classification and 49.00\% on Furniture Classification, 
highlighting the need for further fine-tuning; %the LMM for the target downstream tasks; 
(2) Our error-driven tuning framework significantly improves performance across different training scales, with a notable 10.3\% average boost across seven tasks at the 100K tuning sample scale compared to the pre-trained LMM;
%With our error-driven model tuning framework, the LMM's performance can be significantly improved across different training scales. Notably, our approach achieves an average performance boost of 7.01\% across seven tasks on the 100K tuning sample setting, compared to the pre-trained LMM. 
(3) By carefully analyzing the missing skills of the pre-trained LMM, our approach is consistently more effective at adapting it to the target task than other baselines across different training scales; %\hl{For example, our method surpasses the Random Sampling baseline by up to 14.5\% in the Image-Caption Matching task and achieves an average improvement of 4.56\% across seven tasks at the 10K training scale.
%more efficient in adapting it to the downstream tasks than the other baseline approaches 
 (4) Random Sampling shows comparable performance as other baselines, which is consistent with previous studies~\cite{Mengzhou_2024,chen2024your} and might be attributed to the positive effects of data diversity. However, it's unstable and sometimes results in poorer performance, e.g., 10K training scale for Image-caption Match; (5) Superfiltering performs the worst among baselines, as GPT-2's inability to process image input limits its performance; 
%Superfiltering delivers the poorest performance among the baselines because the selector model, GPT-2, cannot process image input. This limitation hinders its ability to accurately assess instruction-following difficulty, leading to poor data quality and diversity. 
(6) Remarkably, using just 6\% of the full supporting dataset (100K samples), our approach achieves at least 94.57\% of the \textbf{Full Data} performance across all benchmarks and even outperforms the \textbf{Full Data} setting on five tasks, indicating that training LMMs with large-scale task-agnostic datasets may suffer from task interference issue~\cite{wang2023far,shen2024multimodal} and hinder the development of task-specific capabilities, highlighting the necessity of targeted data selection for more efficient model adaptation; %generic LMMs to specific downstream tasks. 
% hinder the development of some specific capabilities and decrease performances on specific tasks~\cite{wang2023far,shen2024multimodal}, and 
% emphasizes the necessity of our proposed targeted data selection method for more efficiently adapting generic LMM to specific downstream tasks. 
(7) More complex tasks usually require more training samples, e.g., 
%The more complex the tasks, the more training samples are required, e.g., 
%require more training samples for a performance boost, e.g., the performance of the 
Image-Caption-Match and Living Thing Classification can be significantly improved by our approach with 10K training samples while the VQA task requires 100K. (8) Based on the main results presented in Table~\ref{tab:main_experiment_llava} and the supplementary analysis in Table~\ref{tab:performance_gap} (Appendix~\ref{sec:performance_variations}), we observe that tasks involving fewer reasoning steps typically achieve greater performance improvements, whereas tasks with longer reasoning chains exhibit comparatively limited gains. We attribute this pattern to two primary factors: (a) Error Localization Complexity—the challenge of accurately identifying the erroneous reasoning step intensifies as reasoning chains grow longer; and (b) Inherent Task Difficulty—tasks requiring longer reasoning chains are inherently more complex, thus making them more challenging targets for performance enhancement.

\begin{table}[hbt!]%{r}{0.45\textwidth}%[h]
\small
\centering
\resizebox{0.5\textwidth}{!}{%
\begin{tabular}{l|c|c|c| c }
\toprule

\multirow{2}{29pt}{\textbf{Method}}&\textbf{\# of Validation} &\textbf{\# of Tuning}    & \textbf{Furniture}  & \textbf{Image-Cap}     \\ 
& \textbf{Samples} & \textbf{Samples}  & \textbf{CIs}& \textbf{Match} \\

% \textbf{Method}&\textbf{\# of Validation Samples}&\textbf{\# of Tuning Samples}  &\textbf{Furniture Cls} &\textbf{Image-Cap Match}   \\ % &

\midrule
Pre-trained &-& -& 49.00  & 64.10  \\ %300
\midrule

Our Approach &0.1K &10K   &  62.10    & 75.90 \\ %404,408
Our Approach &1K &10K  &64.80   &77.70  \\
\midrule

Our Approach &0.1K &100K   & 67.00   &78.50  \\ %404,408
Our Approach &1K &100K  &67.70  & 80.10 \\

\bottomrule
\end{tabular}%
}
\caption{Experiments with different sizes of validation set, with LLaVA-7B as the student model and GPT-4o-mini as the teacher model.%\lifu{split the long name, e.g., \# of Validation Samples into two rows}\barry{Done}
}
\label{tab:validation_set_scale}
\vspace{-3mm}
\end{table}

\paragraph{Requirement of a Small Validation Set} 
While our error-driven, data-efficient tuning framework shows significant improvements on various downstream tasks, we acknowledge that the need for a validation set for each target task could limit generalizability. However, our approach only requires a small validation set—around 1K samples—which is more feasible than large, human-annotated, task-specific training datasets. To reduce this cost, we tested using a smaller validation set. As shown in Table~\ref{tab:validation_set_scale}, even with just 100 validation samples, our framework still enhances pre-trained LMMs, achieving a good cost-performance balance. However, performance slightly decreases compared to the 1K-sample setting, likely due to reduced diversity and skill coverage. Future work could explore using closed-source LMMs to generate pseudo-answers for an unlabeled validation set, reducing the need for manual labeling.

\begin{table*}[hbt!]%{r}{0.45\textwidth}%[h]
\small
\centering
\resizebox{1\textwidth}{!}{%
\begin{tabular}{l|c|c|c| c|c|c|c|c|c}
\toprule
\textbf{Method} &\textbf{Teacher} &\textbf{\# of Tuning Samples} &\textbf{MM-Bench}    &\textbf{Appliance Cls}  &\textbf{Furniture Cls} &\textbf{Living Thing Cls} &\textbf{VQA}&\textbf{Image-Cap Match}    &\textbf{ScienceQA}  \\ % &

\midrule
Pre-trained LMM &- & 0  & 64.30 &45.80 &  49.00&  79.40& 77.00&  64.10 &65.34\\ %300 

\midrule
 
Our Approach &LLaVA-72B &10K  & 63.55 &  62.00  & 64.40& 89.00 & 75.50  & 75.00 & 64.90\\  %
Our Approach&LLaMA-90B&10K & 62.95 &61.90 & 	64.50 & 89.20&75.50 & 76.40&65.89 \\
Our Approach &GPT-4o-mini &10K   &  63.86  &  62.10&  64.80  &   90.60  &   76.00   &  77.70  & 65.89   \\  % 
Our Approach&GPT-4o&10K & 64.01 &62.50   & 65.10 &91.60 &76.30 &	77.90&65.94\\

\midrule 
Our Approach&LLaVA-72B &100K  & 63.25 & 63.40 & 67.00   & 93.20 &  77.60  & 78.60 & 66.58 \\  
Our Approach&LLaMA-90B&100K & 64.16 &64.00 & 67.50  & 93.30 &77.80 & 	78.70&66.63 \\
Our Approach&GPT-4o-mini &100K &   64.46  &  64.10 &  67.70  &  93.60  &  79.00    &  80.10    &  68.02   \\  %   

 Our Approach &GPT-4o&100K& 64.46 &64.80 & 68.00 & 93.60&79.10 & 	81.10&68.07\\

\bottomrule
\end{tabular}%
}
\vspace{-3mm}
\caption{Evaluation results when using LLaVA-v1.5-7B as the student model and LLaVA-OneVision-72B, LLaMA-90B, GPT-4o-mini, and GPT-4o as different teacher models.  
}
\label{tab:choice_of_teacher}
\vspace{-2mm}
\end{table*}

\begin{table*}[hbt!]%{r}{0.45\textwidth}%[h]
\small
\centering
\resizebox{1\textwidth}{!}{%
\begin{tabular}{l|c |c| c|c|c|c|c|c}
\toprule
\textbf{Method}&\textbf{\# of Tuning Samples}     &\textbf{MM-Bench}   &\textbf{Appliance Cls}  &\textbf{Furniture Cls} &\textbf{Living Thing Cls} &\textbf{VQA}&\textbf{Image-Cap Match}   &\textbf{ScienceQA}  \\ % &

\midrule
Pre-trained LMM & 0  & 82.80$^\dagger$   & 63.70 &67.60  &  93.60 &  87.90&  84.30  & 85.50$^\dagger$ \\
\midrule
 Our Approach&10K  &  82.08 & 64.60 & 69.90 &  94.00    &   88.30   &  88.00  &  85.47 \\  % 
Our Approach&100K  & \textbf{82.83} & \textbf{66.20} & \textbf{71.40} &  \textbf{95.80}  &  \textbf{88.50}   & \textbf{88.60}     &\textbf{87.36}\\  %   

\bottomrule
\end{tabular}%
}
\vspace{-3mm}
\caption{Evaluation results when using Qwen2-VL-7B as the student model and GPT-4o-mini as the teacher model. $^\dagger$Reported in previous work. %\textbf{Bold} numbers denotes the best performing setting for each task.
}
\label{tab:different_student}
\vspace{-5mm}
\end{table*}

\paragraph{Results of Different Student and Teacher Models}
To demonstrate the generalizability of our framework, we employ different LMMs as student models or teacher models and show the performance on seven downstream tasks. Specifically, Table~\ref{tab:choice_of_teacher} shows the performance of our framework when utilizing LLaVA-v1.5-7B as the student model, and LLaVA-72B, LLaMA-3.2-90B-Vision~\cite{grattafiori2024llama}\footnote{https://huggingface.co/meta-llama/Llama-3.2-90B-Vision-Instruct}, GPT-4o-mini, or GPT-4o~\cite{achiam2023gpt} (\texttt{gpt-4o-2024-11-20})\footnote{GPT-4o is approximately 16.7 times more expensive than GPT-4o-mini, which is why we opted not to use it in our main experiments.}  as the teacher model. Despite the capability gap between these teacher models on general multimodal tasks, their performance is quite comparable when utilizing them as the teacher model in our framework, demonstrating the generalizability and robustness of our framework.\footnote{We further discuss the effectiveness of these teacher models in skill analysis in Appendix~\ref{sec:teacher_model_comparison}.}  Additionally, Table~\ref{tab:different_student} shows the performance of our framework when using Qwen2-VL-7B as the student model and GPT-4o-mini as the teacher model. Note that Qwen2-VL-7B was the state-of-the-art LMM under 10B parameters at the time of submission.
As we can see, though the pre-trained Qwen2-VL-7B has already significantly outperformed LLaVA-v1.5-7B across
all downstream tasks, by employing our error-driven data-efficient tuning framework, its performance can be further improved by up to 4.30\%, which further underscores the potential of our framework for effectively adapting state-of-the-art generic LLMs to specific downstream tasks.

\subsection{Ablation Study}

\begin{table}[hbt!]%{r}{0.45\textwidth}%[h]
\small
\centering
\resizebox{0.48\textwidth}{!}{%
\begin{tabular}{l|c|c|c }
\toprule
\multirow{2}{29pt}{\textbf{Method}}&\textbf{\# of Tuning}    & \textbf{Furniture}  & \textbf{Image-Cap}     \\ 
& \textbf{Samples}  & \textbf{CIs}& \textbf{Match} \\
% &

\midrule
Pre-trained LMM & 0& 49.00  & 64.10  \\ 
\midrule
\rowcolor{lightgray}
\textbf{Ours} & 10K  & \textbf{64.80} & \textbf{77.70}\\
Ours w/o Mistake Identification&10K   &   63.90   & 73.00\\ 
Ours w/o Error Collection&10K   &    63.10 &70.10 \\ 
Ours w/o Skill Analysis&10K  &   62.30  & 69.80 \\
Ours w/o Targeted Tuning&10K  & 60.60 & 65.03 \\
 \midrule
 \rowcolor{lightgray}
\textbf{Ours} &30K  & \textbf{67.10} & \textbf{80.00}\\
Ours w/o Mistake Identification&30K   &  65.90 & 78.30\\
Ours w/o Error Collection&30K   &   65.60  &77.00 \\

Ours w/o Skill Analysis&30K  &  64.70   &74.30  \\
Ours w/o Targeted Tuning&30K  & 63.13 & 71.97 \\
\bottomrule
\end{tabular}%
}
\vspace{-2mm}
\caption{Ablation study with LLaVA-v1.5-7B as the student model and GPT-4o-mini as teacher model. (\%)}
\label{tab:ablation_study}
\vspace{-4mm}
\end{table}

As shown in Table~\ref{tab:ablation_study}, we conduct ablation studies to demonstrate the effectiveness of each key component in our framework, using LLaVA-v1.5-7B as the student model, GPT-4o-mini as the teacher model, and Furniture Classification and Image Caption Match as the downstream tasks. 
We can see that: (1) Without the \textbf{Mistake Identification} module, performance drops by up to 4.70\%, highlighting the challenge of directly analyzing missing skills from lengthy rationales; (2) Extracting skills from the entire validation dataset rather than error samples (i.e., \textbf{w/o Error Collection}), leads to a 7.6\% performance drop at the 10K training scale, indicating inefficiency with limited training resources; 
%when we conduct skill analysis on error samples without leveraging our \textbf{Mistake Identification} module to identify the true mistake steps (\textbf{Ours w/o Mistake Identification}), the performance drops up to 4.70\%. This decline may be attributed to the challenge of directly analyzing missing skills within the lengthy rationales provided by the student LMM; 
%\hl{(2) When we extract skills directly from the entire validation dataset instead of from student LLM's error samples (\textbf{Ours w/o Error Collection}), the performance drops 7.6\% in 10K training scale setting, which indicates its inefficiency in addressing LMM’s deficiencies with a limited training budget.}\barry{review R1W1} 
(3) Using mistake steps as queries for targeted training samples retrieval (i.e., \textbf{w/o Skill Analysis}) results in a 7.90\% performance drop, which is 
%(3) when directly leveraging the mistake step as the query to retrieve targeted training samples from the supporting dataset (\textbf{Ours w/o Skill Analysis}), we can observe a significant performance drop up to 7.90\%. This performance drop is 
expected since the query used for data retrieval (i.e., mistake step) is not precisely aligned with the index of the supporting dataset (i.e., skills), though there is a correlation between them; 
(4) Randomly sampling from the supporting dataset (\textbf{w/o Targeted Tuning}) also leads to consistent performance drops, confirming the importance of error-driven data selection for effective tuning.

\subsection{Cost-Benefit Analysis}\label{sec:cost_benefit}

\begin{table*}[hbt!]%{r}{0.45\textwidth}%[h]
\small
\centering
\resizebox{1\textwidth}{!}{%
\begin{tabular}{l|c|c|c|c|c|c| c|c|c|c }
\toprule
\textbf{Method}&\textbf{Teacher}&\textbf{Task}&\textbf{$\mathcal{D}_{val}$ } &\textbf{Mistake+Skill (s)}&\textbf{Data Retrieval (s)}  &\textbf{Fine-tuning (s)}&\textbf{$\mathcal{D}_{train}$}  &\textbf{Accuracy (\%)} &\textbf{Complexity}&\textbf{API cost (\$)}    \\ % &

\midrule
Pre-trained &- & ScienceQA &1K&   -& - & - &- & 65.34 &-&0\\
Full Data &- & ScienceQA&1K& - & - &72,410 &1,552K&67.23 &$\mathcal{O}(|\mathcal{D}_{support}|)$ &0 \\ 

Ours&LLaVA-72B &ScienceQA&1K&224 & 50 &4,353  &100K&  66.58  & $\mathcal{O}(|\mathcal{D}_{train}|+|\mathcal{D}_{error}|*t)$&0  \\
Ours&GPT-4o-mini& ScienceQA &0.1K &29 & 7 & 4,647 &100K & 66.83  & $\mathcal{O}(|\mathcal{D}_{train}|+|\mathcal{D}_{error}|*t)$ & 0.1  \\
Ours&GPT-4o-mini& ScienceQA &1K &219 & 54 &4,041&100K & 68.02 & $\mathcal{O}(|\mathcal{D}_{train}|+|\mathcal{D}_{error}|*t)$ & 1.1  \\
\midrule
Pre-trained &- & Image-Cap Match&1K&   -& - & - &- & 64.10 &-&0\\

Full Data &- &Image-Cap Match&1K&   -   &-&72,410 &1,552K&84.70 &$\mathcal{O}(|\mathcal{D}_{support}|)$ &0\\ 

Ours&LLaVA-72B &Image-Cap Match&1K&177 &  48&  4,534&100K&  78.60  & $\mathcal{O}(|\mathcal{D}_{train}|+|\mathcal{D}_{error}|*t)$&0  \\
Ours&GPT-4o-mini &Image-Cap Match&0.1K&  27&    6& 4,296  &100K&   78.50 & $\mathcal{O}(|\mathcal{D}_{train}|+|\mathcal{D}_{error}|*t)$ & 0.1 \\
Ours&GPT-4o-mini &Image-Cap Match&1K&198 & 50  &  4,391&100K&  80.10  & $\mathcal{O}(|\mathcal{D}_{train}|+|\mathcal{D}_{error}|*t)$ &0.8  \\

 % Random &- & ScienceQA& - &  - &4,500&100K &65.74 &$\mathcal{O}(|\mathcal{D}_{train}|)$&0 \\
 % Random &- & Image-Cap Match&   -& - & 4,300 &100K & 78.30 &$\mathcal{O}(|\mathcal{D}_{train}|)$&0\\
 % Random &- & Image-Cap Match&   -& - & 4,500 &100K & 78.30 &$\mathcal{O}(|\mathcal{D}_{train}|)$&0\\
\bottomrule
\end{tabular}%
}
\caption{Cost-benefit analysis on ScienceQA and Image-caption Matching tasks. $\|D_{support}\|$, $\|D_{train}\|$, $\|D_{val}\|$, $\|D_{error}\|$, and $t$ represent the sample size of supporting dataset, training dataset, validation dataset, error set, and average reasoning steps, respectively. \textbf{Mistake+Skill}, \textbf{Data Retrieval}, and \textbf{Fine-tuning} indicate the runtime (in seconds) for Mistake Identification and Skill Analysis, Training Data Retrieval, and Fine-tuning stages, respectively, measured on $8\times40$GB A100 GPUs. \textbf{API cost} represents the cost (in USD)  for closed-source LMM.}
\label{tab:cost_benefit}
\vspace{-3mm}
\end{table*}

We perform a cost-benefit analysis using the ScienceQA and Image-Caption Matching tasks. Table~\ref{tab:cost_benefit} demonstrates that our framework incurs significantly lower overhead compared to the \textbf{Full Data} setting, leading to substantial reductions in tuning costs. To further optimize the cost-performance balance, the size of the validation set can be reduced (e.g., to 0.1K samples), accelerating the tuning process. Additionally, adopting open-source LLMs instead of proprietary API-based models can eliminate associated monetary costs while still maintaining robust performance gains over baseline methods.

\subsection{Effectiveness of Mistake Identification}
\label{sec:mistake_identification}
We further evaluate the effectiveness of our \textbf{Mistake Identification} method and compare it with three baselines: (1) \textbf{Random}, where an intermediate step is randomly selected as the mistake step;  (2) \textbf{Prompt Per Step}~\cite{tyen2024llms}, where GPT-4o-mini is prompted to verify the correctness of each intermediate reasoning step, selecting the first incorrect one as the mistake step; (3) \textbf{Pseudo Rationale Match}, where GPT-4o-mini is first prompted to generate a sequence of pseudo reasoning steps based on the question and gold answer and compare them with the reasoning steps generated by the student model to find the mistake step. Due to the lack of gold labels for mistake steps in the validation datasets, 
we sample 100 error samples from the validation set of ScienceQA and manually label the mistake step for each error sample. The annotation process is detailed in Appendix~\ref{sec:annotation}.

\begin{table}[hbt!]%{r}{0.45\textwidth}%[h]
\small
\centering
\resizebox{0.5\textwidth}{!}{%
\begin{tabular}{l|c|c  }
\toprule
  \textbf{Method}&\textbf{Accuracy (\%)} &\textbf{Recall@3}   \\ % &
\midrule

 Random  &7.0 &  16.0 \\
Prompt Per Step~\cite{tyen2024llms}      & 28.0 &34.0 \\
Pseudo Rationale Match   &59.0 &  68.0  \\ 
\rowcolor{lightgray}
Our Method   &\textbf{65.0}  & \textbf{77.0} \\ %, 0.5, 10, 4-1
\bottomrule
\end{tabular}%
}
\vspace{-2mm}
\caption{Evaluation of various mistake identification methods on ScienceQA. Recall@3 quantifies the percentage of evaluation samples where the annotated gold mistake step matches the predicted step or falls within the three preceding steps.
}
\label{tab:mistake_identification_scienceqa}
\vspace{-3mm}
\end{table}

As shown in Table~\ref{tab:mistake_identification_scienceqa}, the \textbf{Random} baseline achieves only 7.0\% accuracy, reflecting the difficulty of mistake identification given that there are 15.22 reasoning steps per sample on average in the validation set. %, which is consistent with the fact that, on average, there are 15.22 reasoning steps per sample in the validation set. 
\textbf{Prompt Per Step} outperforms \textbf{Random}, though it tends to incorrectly mark steps as wrong when they cannot be directly inferred from previous ones. 
%. By checking its error cases, we see that the baseline method tends to verify if the current step can be directly inferred from the preceding steps. If not, it marks the current step as incorrect. 
For example, given the following reasoning steps: ``\textit{Magnet sizes affect the magnitude of the magnetic force}. \textit{Imagine magnets that are the same shape and material}. \textit{The larger the magnets, the greater the magnetic force}.'', \textbf{Prompt Per Step} identifies the second step as incorrect because ``\textit{The context doesn't indicate that they are all identical in shape or size. So this rationale step is incorrect.}''. Instead, our mistake identification method surpasses all baselines by effectively analyzing the dynamics of the probabilities for each candidate answer from the teacher model, demonstrating its robustness.

\paragraph{Error Propagation} We conduct additional experiments on ScienceQA to assess the impact of error propagation from the Mistake Identification stage on the model's overall performance. For this, we manually annotate mistake steps and missing skills in 100 error samples. These annotated missing skills are used to retrieve training samples for fine-tuning the student LMM, establishing the \textbf{Gold Mistake Step} setting. In the \textbf{Predicted Mistake Step} setting, we leverage the predicted mistake step from our method with about 65\% accuracy for mistake identification. As shown in Table~\ref{tab:error}, our method can achieve comparable performance with \textbf{Gold Mistake Step} setting, indicating limited error propagation. This effectiveness is likely due to the fact that in 77\% of the samples, as shown in Table~\ref{tab:mistake_identification_scienceqa}, the annotated gold mistake step matches the predicted step or falls within the three preceding steps, allowing our approach to offer hints for identifying the mistake context across all reasoning steps even if the exact mistake step is not identified, thereby enhancing the skill analysis stage. Based on predicted mistake step, we leverage our skill analysis module to generate missing skills and manually verify their effectiveness. We find that in 87\% of the samples, the generated missing skills match with the annotated gold missing skills, as detailed in Appendix~\ref{sec:teacher_model_comparison}. %\lifu{what does this mean?}\barry{added more details}

% \begin{table}[hbt!]%{r}{0.45\textwidth}%[h]
% \small
% \centering
% \resizebox{0.5\textwidth}{!}{%
% \begin{tabular}{l|c|c|c }
% \toprule
% \textbf{Setting}&\textbf{Mistake Identification Accuracy (\%)}&\textbf{\# of Tuning Samples}  &\textbf{ScienceQA}   \\ % &

% % \midrule
% % Pre-trained &-& -& 49.00  & 64.10  \\  
% \midrule

% Gold Mistake Step &100 &10K   &   65.74    \\  
% Predicted Mistake Step&65 &10K  & 65.54  \\
% \midrule

% Gold Mistake Step &100 &30K   &    67.07   \\  
% Predicted Mistake Step &65 &30K  &  66.48  \\

% \bottomrule
% \end{tabular}%
% }
% \caption{Experiments to assess error propagation in our approach.}
% \label{tab:error}
% \vspace{-3mm}
% \end{table}

\begin{table}[hbt!]%{r}{0.45\textwidth}%[h]
\small
\centering
\resizebox{0.47\textwidth}{!}{%
\begin{tabular}{l|c|c }
\toprule
\textbf{Setting}&\textbf{\# of Tuning Samples}  &\textbf{ScienceQA}   \\ % &

% \midrule
% Pre-trained &-& -& 49.00  & 64.10  \\  
\midrule

Gold Mistake Step &10K   &   65.74    \\  
Predicted Mistake Step &10K  & 65.54  \\
\midrule

Gold Mistake Step &30K   &    66.93   \\  
Predicted Mistake Step &30K  &  66.48  \\

\bottomrule
\end{tabular}%
}
\vspace{-2mm}
\caption{Impact of error propagation from mistake identification on the model's overall performance.}
\label{tab:error}
\vspace{-5mm}
\end{table}

% \subsection{Hyperparameter Sensitivity of Mistake Identification}\label{sec:mistake_hyper}\barry{This section is moved from Appendix to main paper}
% 

\begin{table}[hbt!]%{r}{0.45\textwidth}%[h]
\small
\centering
\resizebox{0.5\textwidth}{!}{%
\begin{tabular}{l|c|c|c |c }
\toprule  
\textbf{$\delta$}&\textbf{ $lambda$ }&\textbf{ \# Tuning }&\textbf{MI Accuracy (\%)}  &\textbf{ScienceQA Accuracy (\%)}  \\ % &

\midrule
  0.8& 12  &10K&60& 64.80  \\
   0.2& 0&10K& 62&  65.20 \\
  0 & 12  &10K&  63 &  65.64\\ 
  \rowcolor{lightgray}0.2  &12    &10K&  65& 65.89\\
\midrule
 0.8& 12  &100K&60&66.53\\
 0.2& 0&100K& 62& 66.83\\
  0 & 12  &100K&  63 &67.18 \\
  \rowcolor{lightgray}0.2  &12   &100K&  65& 68.02\\

\bottomrule
\end{tabular}%
}
\caption{Hyperparameter sensitivity analysis of the Mistake Identification (MI) method. \textbf{MI Accuracy} denotes the accuracy of correctly identifying the erroneous step, while \textbf{ScienceQA Accuracy} reflects the downstream task performance after tuning with the selected samples.}
\label{tab:mistake_identification_hyper}
\vspace{-3mm}
\end{table}

\paragraph{Hyperparameter Sensitivity}\label{sec:mistake_hyper}
We evaluate the robustness of the Mistake Identification (MI) method with respect to two key hyperparameters: the minimum probability gap threshold ($\delta$) and the persistence window ($\lambda$), introduced in Section~\ref{sec:mistake_identify}. The optimal values, $\delta = 0.2$ and $\lambda = 12$, are selected via a comprehensive grid search over $\delta \in \{0, 0.1, \ldots, 0.9\}$ and $\lambda \in \{0, 1, \ldots, 19\}$, using accuracy on the annotated validation set (see Appendix~\ref{sec:annotation}) as the selection criterion. As shown in Table~\ref{tab:mistake_identification_hyper}, variations around the optimal values lead to limited declines in both MI accuracy and downstream ScienceQA performance, suggesting that our method maintains practical robustness to changes in hyperparameters.

\section{Conclusion}
We propose a novel error-driven, data-efficient tuning paradigm to effectively adapt generic, pre-trained large multimodal models (LMMs) to various new and emerging downstream tasks without requiring any task-specific training samples. Extensive experiments show that our framework can significantly improve pre-trained LMM's performance on seven downstream tasks by retrieving targeted tuning samples from the supporting dataset. Future work can explore loss-driven latent skills~\cite{Xu_Zifan_2023} to support more fine-grained skills.

\section*{Limitations}
%\paragraph{Continous Instruction Tuning} 
%We discuss the limitations of our approach here. 

Though the extensive experiments have demonstrated the effectiveness of our error-driven data-efficient tuning framework, it still has several limitations: (1) \textbf{Requirement of Validation Set}. The task-specific validation set is crucial in our framework to measure the downstream task distribution and LMM's capability gaps. For certain tasks, even creating and labeling 1,000 samples could be expensive and time-consuming. Further research is necessary to remove the requirement of such task-specific validation sets. (2) \textbf{Mistake Identification Needs Further Improvement}. In this work, we develop a straightforward yet effective method for identifying mistakes within the rationales of LMMs. However, there is still potential for further enhancing this component, which is crucial for precisely analyzing the capability gaps of LMMs for target downstream tasks.

\section*{Ethics Statement}
We carefully follow the ACM Code of Ethics \footnote{\url{https://www.aclweb.org/portal/content/acl-code-ethics}} and have not found potential societal impacts or risks so far. To the best of our knowledge, this work has no notable harmful effects and uses, environmental impact, fairness considerations, privacy considerations, security considerations, or other potential risks.

\section*{Acknowledgements}
This research is partially supported by the award No. 2238940 from the Faculty Early Career Development Program (CAREER) of the National Science Foundation (NSF), the U.S. DARPA ECOLE Program \#HR001122S0052, and FoundSci Program \#HR00112490370. The views and conclusions contained herein are those of the authors and should not be interpreted as necessarily representing the official policies, either expressed or implied, of the U.S. Government. The U.S. Government is authorized to reproduce and distribute reprints for governmental purposes notwithstanding any copyright annotation therein.
%This research is based upon work supported by \barry{funder} Program \# \barry{funding ID}. The views and conclusions contained herein are those of the authors and should not be interpreted as necessarily representing the official policies, either expressed or implied, of \hl{DARPA or the U.S. Government}\barry{funder}. The \hl{U.S. Government} is authorized to reproduce and distribute reprints for governmental purposes notwithstanding any copyright annotation therein. 

\bibliography{anthology,section/custom}
\clearpage
\appendix

% \section{Additional Related Work}\label{sec:related_work}

% \paragraph{Knowledge Distillation} \hl{While our framework includes a teacher model, it serves primarily as an analytical tool to diagnose the student model's deficiencies. It does not transfer knowledge directly but informs the data selection process, ensuring that the training precisely targets and addresses the gaps in the student model's capabilities.}\barry{Review R2S1}

\begin{table*}[hbt!]%{r}{0.45\textwidth}%[h]
\small
\centering
% \resizebox{0.9\textwidth}{!}{%
\begin{tabular}{l|c}
\toprule    
\textbf{Answer Format}  &\textbf{Regular Expression Pattern }   \\ % &
\midrule
Answer is (A) & \text{(?i)answer is \textbackslash (([A-Z])}  \\ 
Answer is (A & \text{(?i)answer is \textbackslash (([A-Z])}  \\ 
Answer is A. & (?i)answer is ([A-Z])\textbackslash. \\
Answer: A & (?i)answer:\textbackslash s?([a-z]) \\
A is the correct answer & (?i)([A-Z]) is the correct \\
A & (?<!\textbackslash S)[a-zA-Z](?!\textbackslash S)(?!.*[a-zA-Z])  \\
answer is the option A & (?<!\textbackslash S)[a-zA-Z] (?!\textbackslash S)(?!.*[a-zA-Z])  \\
choose the answer, A & (?i)choose the answer,\textbackslash s?([a-z]) \\
\bottomrule
\end{tabular}%
% }
\caption{Answer format table}
\label{tab:answer_format}
\vspace{-3mm}
\end{table*}

\section{Answer Format}\label{sec:answer_format}
Table ~\ref{tab:answer_format} shows the answer formats that we use to parse the answer.

\section{Prompt Template}

\subsection{Inference Prompt Template}\label{sec:inference_prompt}
\begin{figure*}[!ht] 
 
    \includegraphics[width=1\textwidth]{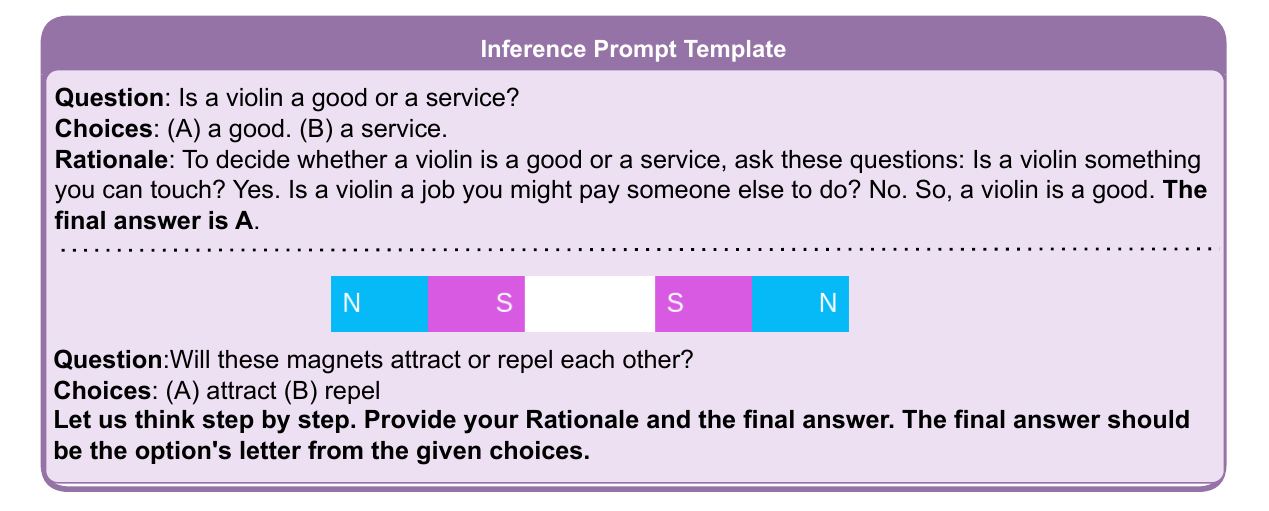}
    \caption{ One example prompt for ScienceQA task to obtain the student model's prediction.}
    \label{fig:inference}
\end{figure*}

Figure~\ref{fig:inference} shows the Inference Prompt Template.

 \subsection{Mistake Identification Prompt Template}\label{sec:mistake_prompt}
\begin{figure*}[hbt!] 
    \centering
    \includegraphics[width=1\textwidth]{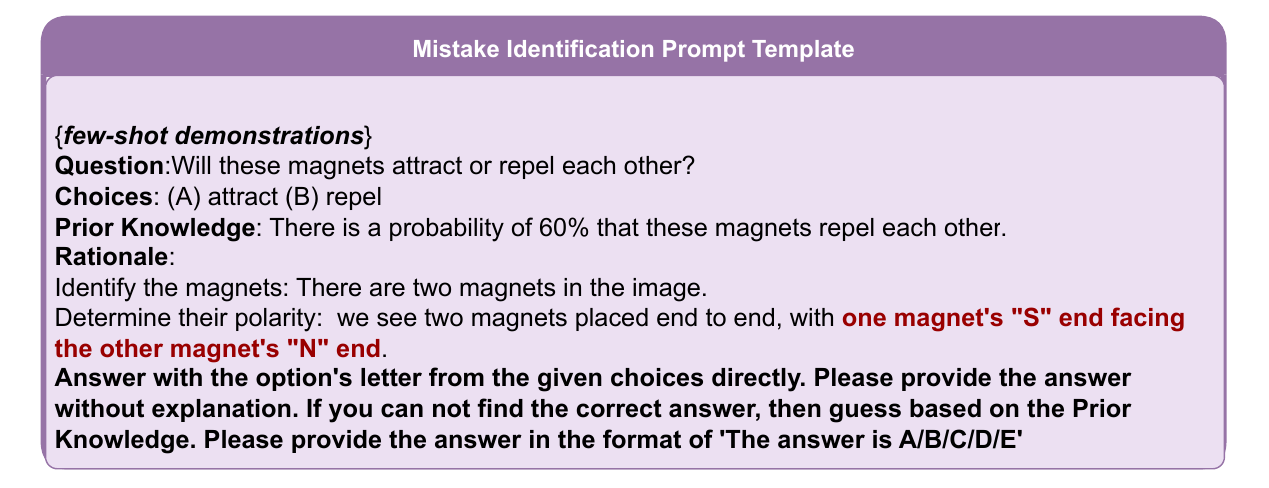}
    \caption{One example prompt to obtain the teacher model's prediction by following the student model's rationale steps. We then identify the mistake rationale step based on the evolution in probabilities of predicted options from the teacher model.}
    \label{fig:mistake_identification_prompt} 
\end{figure*}

 Figure~\ref{fig:mistake_identification_prompt} shows Mistake Identification Prompt Template

 \subsection{Skill Analysis Prompt Template}\label{sec:skill_prompt}
\begin{figure*}[hbt!] 
    \centering
    \includegraphics[width=1\textwidth]{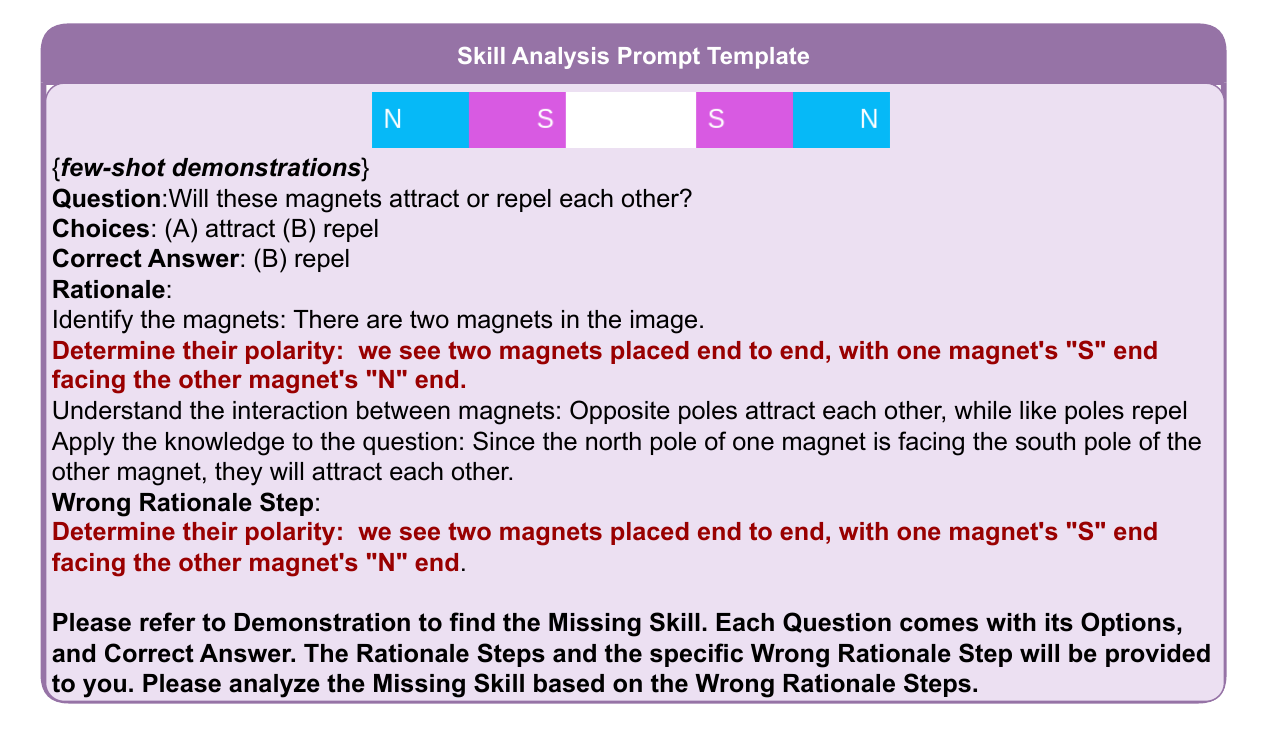}
    \caption{One example prompt to trigger the teacher model to analyze the missing skill based on the wrong rationale step.}
    \label{fig:skill_analysis} 
\end{figure*}

 Figure~\ref{fig:skill_analysis} shows Skill Analysis Prompt Template

 \subsection{Skill Set Analysis Prompt Template}\label{sec:skill_set_prompt}
\begin{figure*}[hbt!] 
    \centering
    \includegraphics[width=1\textwidth]{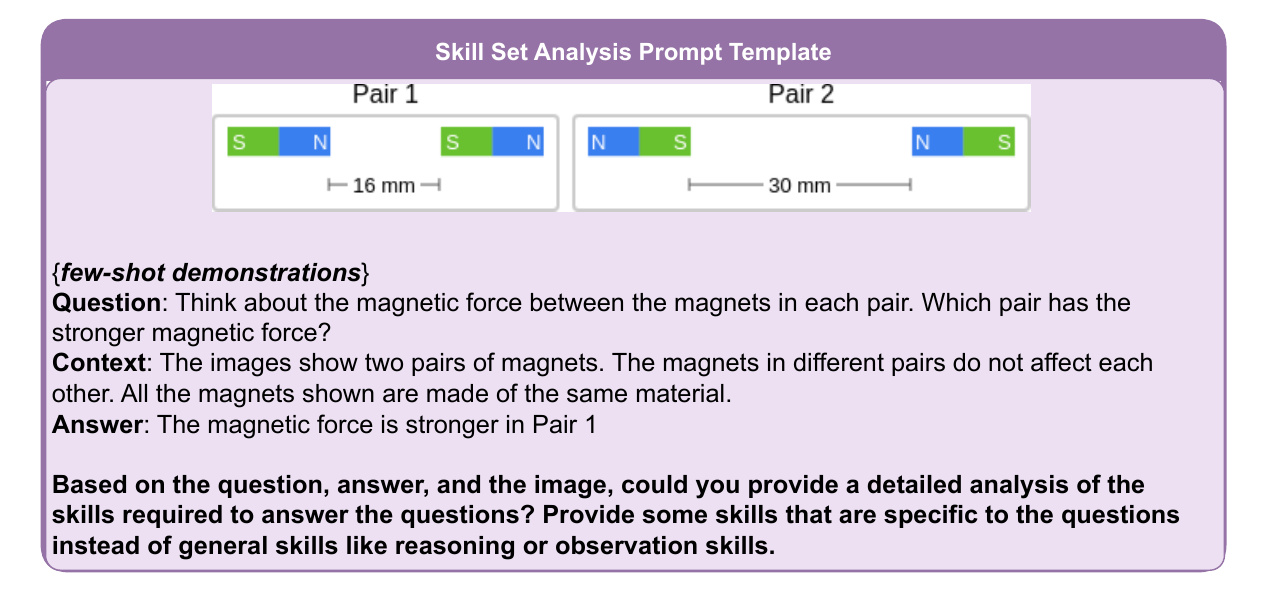}
    \caption{ One example prompt to trigger the teacher model to analyse a sequence of required skills for each sample in the supporting dataset. }
    \label{fig:skill_set_analysis}
\end{figure*}

 Figure~\ref{fig:skill_set_analysis} shows Skill Set Analysis Prompt Template.

\section{Mistake Identification}\label{sec:mistake_hyper_search}

% \subsection{Answer Switch Phenomenon}\label{sec:answer_switch_experiment} 
% Previous work~\newcite{tyen2024llms} shows that replacing the wrong rationale step with a regenerated rationale step may switch LLM's prediction from the wrong answer to the right answer. Inspired by this, we verify the answer-switch phenomenon on our setting. Specifically, we first run the student model on the validation set of ScienceQA dataset and obtain error samples as we mentioned in Section~\ref{sec:error_extraction}. We then randomly select 100 error samples as the validation set for mistake identification method. On this validation set, we monitor the probabilities of the candidate answers from teacher model, as explained in Section~\ref{sec:mistake_identify}. We find that when the teacher model has access to the correct answer as prior knowledge, it initially assigns a higher probability to the correct answer in 89\% of the evaluation samples. As student model's wrong reasoning steps are added, it shifts to the wrong answer in 70\% of samples. Without prior knowledge, the correct answer receives a higher initial probability in 60\% of samples, with a shift to the wrong answer in 43\%.  

\subsection{Human Annotation for Mistake Identification}\label{sec:annotation}
We first run the student model on the validation set of ScienceQA dataset and obtain error samples as we mentioned in Section~\ref{sec:error_extraction}. We then randomly select 100 error samples for annotation. For each error sample, we split the student model's rationale into a sequence of reasoning steps\footnote{Following previous studies~\cite{tyen2024llms}, we treat each sentence in the rationale as one reasoning step.}. The annotator will then annotate these error samples following the following guidelines:
\begin{itemize}  
    \item Open one of your annotation web pages
    \item For each sample, check through the question, the choices, the image, the correct answer, and the wrong prediction. 
    \item Then you will read rationale step one by one and check whether the current rationale step contains logical errors. If yes, you can record the corresponding index (starting from 0).
     \item If you did not record any rationale step after checking all of them, you can provide "-1" as the label of mistake step for this sample.
\end{itemize}

\subsection{Probability Gap Statistic}\label{sec:probability_gap}
\begin{figure}[hbt] %ht!
    % \centering
    \includegraphics[width=0.5\textwidth]{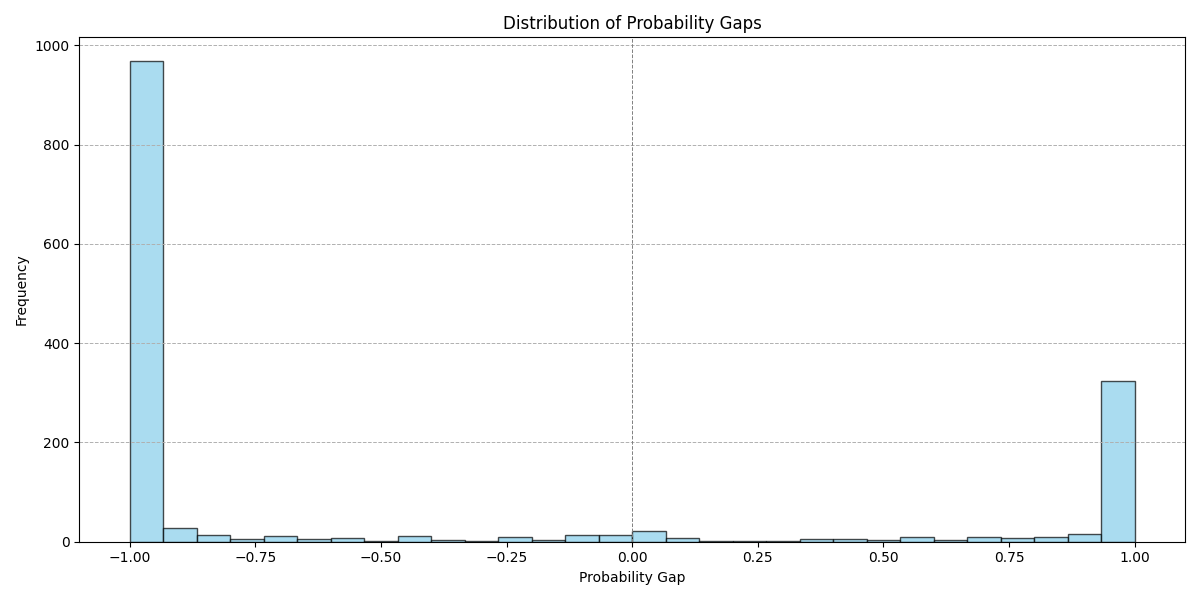}
    \caption{ Probability gap statistics in our annotated validation set, where Probability Gap=p(student model’s wrong answer)-p(correct answer).}
    \label{fig:probability}
\end{figure}

 Figure~\ref{fig:probability} shows probability gap statistics in our annotated validation set, where Probability Gap = p(student model’s wrong answer) - p(correct answer). 

% \subsection{Hyperparameter Search}\label{sec:mistake_real_hyper_search}

% We conduct hyperparameter search for our Mistake Identification method over two hyperparameters using the annotated validation set: (1) $\delta$, the minimum probability gap required between the incorrect and correct answers, and (2) $\lambda$, the minimum number of consecutive reasoning steps for which this gap must persist. These hyperparameters serve to stabilize the detection of mistake steps by mitigating transient fluctuations in the teacher model's output probabilities. The hyperparameter search covers a comprehensive range: $\delta \in {0, 0.1, \ldots, 0.9}$ and $\lambda \in {0, 1, \ldots, 19}$.

\section{Definition and Explanation of skills}\label{sec:skill_definition}
In the education domain, skill is defined as an ability to carry out a task with pre-determined results, often within a given amount of time, energy, or both~\cite{dyatlova2018project}. Some studies stress out the expandability of skill: skill refers to any ability acquired by training or practice, allowing individuals to perform well in multifarious types of tasks~\cite{perez2009understanding,green2011skill}. In this work, we follow ~\cite{10.48550/arxiv.2307.14430} and define a skill $s$ as a unit of behavior with associated data $X$ such that if the LMM is trained on dataset $D$, where $D \subseteq X$, it has improved performance on samples belonging to $X\backslash D$. This definition of a skill is flexible---it focuses on the expandability of skill and means that given a training dataset associated with the skill, a model $f$ has an improved performance when evaluated on validation data associated with this skill. Under this definition, a skill could be a fine-grained, instance-specific ability like ``Identify the poles of a magnet'', instead of general skills like ``color recognition'', ``shape recognition'', and ``texture recognition''.

\section{Experiment Results for Random Sampling}\label{sec:random_sampling}

\begin{table*}[hbt!]%{r}{0.45\textwidth}%[h]
\small
\centering
\resizebox{1\textwidth}{!}{%
\begin{tabular}{l|c|c| c|c|c|c|c|c}
\toprule
\textbf{Method}&\textbf{\# of Tuning Samples} &\textbf{MM-Bench}   &\textbf{Appliance Cls}  &\textbf{Furniture Cls} &\textbf{Living Thing Cls} &\textbf{VQA}&\textbf{Image-Cap Match}  &\textbf{ScienceQA}  \\ % &

\midrule
  
Random 1&10K  &63.40 & 57.70& 61.00& 85.60 &74.80  &63.20  & 64.06     \\ %   
Random 2&10K  &  62.80& 55.30  &  60.30 &  79.10  & 73.00  & 66.30 &63.11 \\ %  
Random 3&10K  & 62.35 &59.40   &60.50  & 81.60   &  74.30 &65.60  & 63.81\\ %  

\midrule

Random 1&30K & 62.65 &61.10 & 63.60& 87.90 & 77.10 & 73.50 & 63.01   \\% 
Random 2&30K& 62.20 & 61.30   & 62.30  & 86.60   &75.40   & 74.10 &63.96 \\ 
Random 3&30K& 62.95 &  60.80  & 63.50  &  86.00  &74.00   &68.30  &63.16 \\

\midrule

Random 1&100K &62.95 &61.20 &66.30 & 91.00 &77.10  &78.30   & 65.74 \\%  
Random 2&100K & 63.86& 62.00  & 66.70&  87.60  &76.30 & 76.40 &  64.55\\% 
Random 3&100K &62.05  &59.30   & 65.70& 87.40   & 77.30&   77.80  &63.36  \\% 

\bottomrule
\end{tabular}%
}
\vspace{-2mm}
\caption{Evaluation results on seven downstream tasks with different numbers of tuning samples retrieved from the supporting dataset. (\%). %\textbf{Bold} numbers denotes the best performing setting for each task on each training data scale. %\lifu{for each setting, highlight the best number for each task}\barry{Should we compute the best accuracy for each training data scale separately or together? If we consider each training data scale separately, then our approach is always the best. If we consider all training data scales together, then sometimes Full Data setting (with 1552K data) achieves the best performance.}\lifu{for each scale}\barry{Done}
}
\label{tab:random_sampling_result}
\vspace{-6mm}
\end{table*}
For the Random Sampling baseline, the random sampling process was repeated three times and we report detailed results for each run in Table~\ref{tab:random_sampling_result}.

\section{Analysis of Task-specific Performance Variations}\label{sec:performance_variations}

\begin{table}[hbt]%{r}{0.45\textwidth}%[h] hbt!
\small
% \centering
\resizebox{0.47\textwidth}{!}{%
\begin{tabular}{l|c|c }
\toprule
\textbf{Task}&\textbf{\# of Reasoning Step}  &\textbf{Performance Gain}   \\ % &

% \midrule
% Pre-trained &-& -& 49.00  & 64.10  \\  
\midrule
Appliance CIs &	7.64	 &18.30 \\  
MM-Bench &	13.23	 &0.16 \\  
Furniture CIs &	8.33 &	18.70 \\  
Living Thing CIs &	6.27	 &14.20 \\  
VQA &	12.74	 &2.00 \\  
Image-Cap Match &	7.13 &	16.00 \\  
ScienceQA &	15.00	 &2.68 \\

\bottomrule
\end{tabular}%
}
\vspace{-2mm}
\caption{Task characteristics for all evaluation tasks. \textbf{\# of Reasoning Steps} indicates the average number of reasoning steps, and \textbf{Performance Gain} refers to the improvement compared with the pre-trained LLaVA-v1.5-7B.}
\label{tab:performance_gap}
\vspace{-5mm}
\end{table}
Table ~\ref{tab:performance_gap} reveals a strong negative correlation between the number of reasoning steps and the performance gains from our framework. We observe that tasks involving fewer reasoning steps typically achieve greater performance improvements, whereas tasks with longer reasoning chains exhibit comparatively limited gains. We attribute this pattern to two primary factors: (a) Error Localization Complexity—the challenge of accurately identifying the erroneous reasoning step intensifies as reasoning chains grow longer; and (b) Inherent Task Difficulty—tasks requiring longer reasoning chains are inherently more complex, thus making them more challenging targets for performance enhancement.

\section{Effectiveness of Different Teacher Models}\label{sec:teacher_model_comparison}
We further discuss whether the choice of the teacher model affects the effectiveness of the skill analysis. We first ask the annotator to write the gold missing skills based on question, reasoning steps, and answer for 100 error samples. We then leverage two teacher models, GPT-4o-mini and LLaVA 72B, to predict missing skills and ask the annotator to manually compare these predicted missing skills with gold missing skills. Our experiments indicate that the missing skills generated by GPT-4o-mini align with the annotated gold missing skills in 87\% of the samples, whereas those generated by LLaVA-72B match the annotated gold missing skills in 79\% of the samples.

\section{Experiment Details}
We conduct experiments on $8\times40$GB A100 GPUs. In the 100K training sample setting, one training can run for around 1.1 hours. We use learning rate as $2 \times 10^{-4}$ and batch size as $128$.

\end{document}